\documentclass[pmlr]{jmlr}

\RequirePackage{graphicx}
 \usepackage{booktabs}
\usepackage{longtable}
\usepackage{wrapfig}

\makeatletter
\def\set@curr@file#1{\def\@curr@file{#1}}
\makeatother
\usepackage[load-configurations=version-1]{siunitx}

\theorembodyfont{\upshape}
\theoremheaderfont{\scshape}
\theorempostheader{:}
\theoremsep{\newline}

\jmlrproceedings{PMLR}{Proceedings of Machine Learning Research}
\jmlrvolume{340}
\jmlryear{2026}
\jmlrworkshop{Machine Learning for Healthcare}

\title[Test-Time Adaptation for EEG Foundation Models]{Test-Time Adaptation for EEG Foundation Models: A Systematic Study under Real-World Distribution Shifts}

\author{\Name{Gabriel Jason Lee$^*$}
        \Email{gjlee4@illinois.edu}
       \AND
       \Name{Jathurshan Pradeepkumar$^*$}
       \Email{jp65@illinois.edu}
       \AND
       \Name{Jimeng Sun}
       \Email{jimeng@illinois.edu}\\
       \addr University of Illinois Urbana-Champaign, Urbana, IL, USA
       }

\usepackage{multirow}
\usepackage{float}

\usepackage{adjustbox}
\usepackage{subcaption}

\begin{document}

\maketitle

\begin{abstract}
Electroencephalography (EEG) foundation models have shown strong potential for learning generalizable representations from large-scale neural data, yet their clinical deployment is hindered by distribution shifts across clinical settings, devices, and populations. Test-time adaptation (TTA) offers a promising solution by enabling models to adapt to unlabeled target data during inference without access to source data, a valuable property in healthcare settings constrained by privacy regulations and limited labeled data. However, its effectiveness for EEG remains largely underexplored. In this work, we introduce \emph{NeuroAdapt-Bench}, a systematic benchmark for evaluating test-time adaptation methods on EEG foundation models under realistic distribution shifts. We evaluate representative TTA approaches from other domains across multiple pretrained foundation models, diverse downstream tasks, and heterogeneous datasets spanning in-distribution, out-of-distribution, and extreme modality shifts (e.g., Ear-EEG). Our results show that the evaluated TTA methods yield inconsistent gains and often degrade performance, with gradient-based approaches particularly prone to heavy degradation and optimization-free methods showing greater stability. For the evaluated EEG foundation models and representative TTA methods, these findings highlight the limitations of directly applying existing TTA techniques to EEG and underscore the need for domain-specific adaptation strategies.

Code is available at \url{https://github.com/leegabriel/NeuroAdapt-Bench}.

\end{abstract}

\section{Introduction}

Electroencephalography (EEGs) offer high-resolution measurements of neuronal activity, capturing brain dynamics at the millisecond scale, making them essential for a wide range of clinical applications, including sleep staging~\citep{phan2022sleeptransformer,pradeepkumar2024toward} and epilepsy diagnosis~\citep{sundaram1999eeg,jia2026odebrain}. Recent advances in self-supervised learning have led to the development of EEG foundation models~\citep{ouahidi2025reve,wang2025cbramod}, which are large neural networks trained on diverse, large-scale EEG corpora to learn generalizable representations. Despite their success, a key barrier to clinical deployment remains: \emph{distribution shift}, where models trained on a given dataset often fail to generalize to new hospitals, acquisition devices, or patient populations.

\begin{wrapfigure}{r}{0.41\textwidth}
    \centering
    \includegraphics[width=0.41\textwidth]{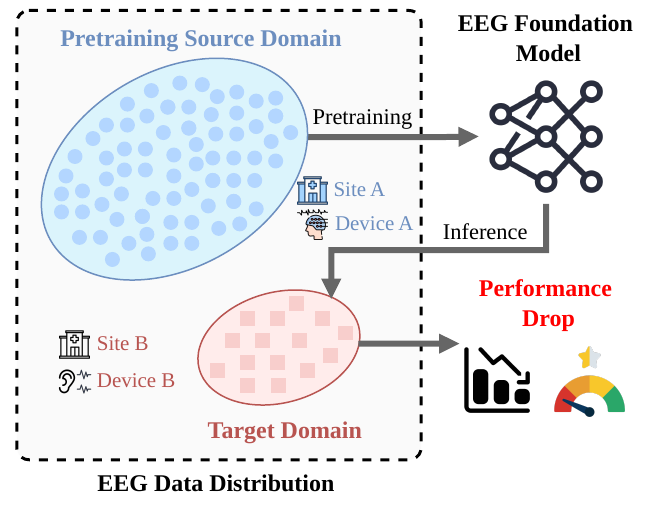}
    \vspace{-0.7cm}
    \caption{Distribution shift in EEG foundation model deployment. Pretrained EEG models often degrade when applied to new sites and devices, motivating the need for label and source-free test-time adaptation.}
    \vspace{-0.2cm}

    \label{fig:overview}
\end{wrapfigure}

As illustrated in Figure~\ref{fig:overview}, distribution shifts are especially severe in EEG analysis. Unlike natural images, where domain gaps are often stylistic, EEG signals exhibit complex, patient-specific dynamics and diverse acquisition protocols that vary substantially across sessions, tasks, and clinical sites~\citep{jayaram2016transfer,yang2023manydg}.
For example,~\citep{kastrati2025eeg} reports substantial performance degradation of EEG foundation models on out-of-distribution tasks such as sleep staging.

Test-time adaptation (TTA) offers a promising solution by enabling models to adapt to target-domain data without labeled samples or access to source data, unlike traditional domain adaptation. This source-free property is particularly valuable in healthcare, where access to source data is often restricted by privacy regulations, limited labeled data, and the computational overhead of model fine-tuning. Prior TTA work in computer vision and speech has introduced a range of strategies, including entropy minimization, continual self-training, prototype adjustment, and source-free pseudo-label refinement~\citep{wang2021tent,Wang_2022_CVPR,iwasawa2021testtime,liang2020we,liu-etal-2024-advancing,wang-etal-2025-dynamic}.

Despite growing interest in TTA across computer vision and speech recognition, its application to EEG remains underexplored. Existing studies typically focus on a single task or architecture, such as driver drowsiness detection or multimodal sleep staging~\citep{jang2025eegtta,guo2025sleeptta,jia2024atta}, which makes it difficult to tell whether observed gains generalize across settings. At the same time, EEG foundation models are explicitly motivated by transfer across datasets and downstream tasks, yet there is still little evidence on how standard TTA methods behave when these models are deployed under realistic EEG distribution shifts. This leaves an important practical gap between pretrained EEG representation learning and reliable deployment.

To address this gap, we conduct a systematic benchmark of test-time adaptation for EEG foundation models. We evaluate representative TTA methods across multiple pretrained EEG foundation models, diverse downstream datasets, and heterogeneous deployment settings, encompassing a range of distribution shifts and tasks, including event detection, abnormality screening, seizure detection, and sleep staging. Overall our contributions are summarized as follows:
\begin{itemize}
    \item \textbf{NeuroAdapt-Bench:} We introduce NeuroAdapt-Bench, a unified benchmark for evaluating test-time adaptation methods on EEG foundation models under realistic distribution shifts.

    \item \textbf{Comprehensive experiments under diverse distribution shifts:} We systematically evaluate TTA methods across a range of EEG tasks and deployment scenarios, including both online and offline adaptation regimes. Our experiments explicitly cover (1) \emph{in-distribution} settings capturing subject-level variability, (2) \emph{out-of-distribution} cross-dataset shifts involving changes in tasks, populations, and acquisition protocols, and (3) \emph{extreme distribution shifts} arising from unseen modalities and recording configurations (e.g., Ear-EEG~\cite{ds005178:1.0.0}). We quantify performance relative to a No-TTA baseline across all settings.

    \item \textbf{Key insights on TTA for EEG:}
    Across the evaluated foundation models and TTA methods, we show that the evaluated TTA methods yield inconsistent gains and often degrade performance under distribution shift. We find that evaluated optimization-free methods (e.g., prototype-based approaches) are generally more stable than gradient-based alternatives, highlighting stability as a central consideration for deployment. In particular, T3A is the only method with a positive mean balanced-accuracy improvement across the in-distribution, out-of-distribution, and extreme-shift Ear-EEG settings. Its largest mean gain is a +18.9 percentage-point improvement in balanced accuracy for REVE-Base on CHB-MIT~\citep{Shoeb2010ApplicationOM}. Detailed per-dataset deltas, relative changes, and statistical significance tests are provided in Appendices~\ref{sec:appendix_delta_tables},\ref{sec:appendix_percentage}, and \ref{sec_appendix_significance}.

    \item \textbf{Open-source benchmark framework:} We release code and evaluation pipelines to facilitate benchmarking of future EEG foundation models and TTA methods. The benchmark will be integrated into an existing Python library to facilitate reproducibility and future research in this domain.

\end{itemize}

\subsection*{Generalizable Insights about Machine Learning in the Context of Healthcare}

Our study highlights several generalizable insights for deploying machine learning in healthcare. First, methods such as test-time adaptation that perform well in domains like computer vision do not necessarily transfer well to EEG signals, where they can introduce instability and degrade performance under realistic distribution shifts. Second, stability and robustness are as critical as accuracy for clinical deployment, and the evaluated optimization-free approaches are more reliable than gradient-based methods in our benchmark. Third, the type and severity of distribution shift, ranging from subject variability to cross-dataset and modality-level differences, strongly impact model performance, underscoring the need for evaluation under realistic conditions. Finally, standardized and reproducible benchmarking frameworks are essential for identifying failure modes and guiding the development of reliable healthcare AI systems.

\section{Related Work}

\subsection{EEG Foundation Models}

Recent advances in large-scale self-supervised pretraining have driven the rapid development of EEG foundation models. These models are motivated by the need to generalize across heterogeneous EEG settings, including differences in subjects, channel configurations, acquisition protocols, and task definitions. However, recent reviews emphasize that current EEG foundation models remain highly heterogeneous in their pretraining data, architectures, and evaluation, leaving their robustness under realistic deployment shift only partially understood~\citep{yao2025eegfm,kuruppu2026eegfmreview}.

Broadly, existing EEG foundation models can be categorized into encoder-only and generative models. Encoder-only models, including BIOT~\citep{yang2023biot}, LaBraM~\citep{jiang2024large}, CBraMod~\citep{wang2024cbramod}, REVE~\citep{ouahidi2025reve}, EEGPT~\citep{wang2024eegpt}, and TFM-Tokenizer~\citep{pradeepkumar2026tokenizing} are primarily optimized for discriminative tasks such as classification. In contrast, generative EEG foundation models~\citep{pradeepkumar2026neural,xu2026sleeplm} focus on language alignment and generative objectives. In this work, we benchmark TTA methods on encoder-based models.

\subsection{Test-Time Adaptation}

Test-time adaptation considers the setting in which a model trained on labeled source-domain data is deployed on unlabeled target-domain data drawn from a shifted distribution. Since deployment-time shift can substantially degrade performance, TTA aims to adapt
the source-trained model during inference using only target samples available at test time, often without access to source data or
target labels~\citep{wang2025otta}. Prior work in computer vision has established several common TTA families, including entropy
minimization, continual self-training, prototype-based adjustment, and source-free pseudo-label
refinement~\citep{wang2021tent,niu2022efficient,Wang_2022_CVPR,iwasawa2021testtime,liang2020we}. Similar approaches have recently emerged in
speech and audio applications under noisy and mismatched deployment conditions~\citep{lin-etal-2024-continual,liu-etal-2024-advancing,wang-etal-2025-dynamic,dong2025ebats}.

Despite progress, TTA for biosignals remains relatively limited and largely task-specific. Recent works have explored approaches such as personalized calibration, teacher–student adaptation, and memory-based stabilization in applications including sleep staging, rPPG, and ECG~\citep{jo2025ttc,guo2025sleeptta,jia2024atta,huang2026rppgtta,wu2026ecgtta}. EEG is particularly challenging in this context due to
its highly non-stationary nature across subjects and sessions, weakly structured relative to signals such as ECG, and sensitivity to
artifacts and acquisition variability~\citep{raj2025eegdenoisingreview}. While initial EEG-specific TTA studies show promising gains in narrowly defined tasks, such as driver drowsiness classification~\citep{jang2025eegtta}, they do not provide a comprehensive understanding of how standard TTA methods generalize across EEG foundation models and downstream tasks.

Our work bridges two emerging directions: EEG foundation models and test-time adaptation. Here, we systematically benchmark representative TTA approaches across multiple EEG foundation models under diverse downstream settings. This enables us to assess not only when adaptation improves performance, but also when it fails, which methods are most stable, and the implications for clinical deployment.

\section{NeuroAdapt-Bench}

\begin{wraptable}{r}{0.4\textwidth}
  \centering
  \vspace{-0.5cm}
  \caption{Classification of the evaluated TTA methods by adaptation regime and update mechanism.}
  \label{tab:tta-methods}

  \resizebox{\linewidth}{!}{
  \begin{tabular}{lccc}
    \toprule
    \textbf{Method}
      & \textbf{Online}
      & \textbf{Batch}
      & \textbf{Gradient} \\
      \textbf{}
      & \textbf{}
      & \textbf{Adaptation}
      & \textbf{Based} \\

    \midrule
    Tent & \checkmark &  & \checkmark \\
    T3A  & \checkmark &  &  \\
    SHOT &  & \checkmark & \checkmark \\
    \bottomrule
  \end{tabular}
  }
   \vspace{-0.5cm}
\end{wraptable}
This section introduces \emph{NeuroAdapt}, our benchmark for systematically evaluating representative TTA methods on EEG foundation models across diverse tasks. Section~\ref{sec:Preliminary} presents the problem formulation and describes the TTA methods considered. Sections~\ref{sec:bench_design} and~\ref{sec:eval_setup} detail the benchmark design and experiment setups.

\subsection{Preliminary}
\label{sec:Preliminary}

In test-time adaptation, a model trained on labeled source-domain data is deployed on unlabeled target-domain data whose distribution may differ from that of the source domain~\citep{pmlr-v119-sun20b}. The objective is to adapt the source-trained model using only unlabeled target samples observed during inference. In this work, we consider three representative methods for this setting (Table~\ref{tab:tta-methods}): Tent~\citep{wang2021tent}, SHOT~\citep{liang2020we}, and T3A~\citep{iwasawa2021testtime}.\\

Under a unified formulation, we represent a source-trained classifier as:
\begin{equation}
f_\theta(x) = h_w(g_\phi(x))
\end{equation}

where $g_\phi$ denotes the feature extractor parameterized by $\phi$, $h_w$ is the classifier head parameterized by $w$, and $\theta = (\phi, w)$ represents the full set of model parameters. Given an input $x$, the model outputs logits $f_\theta(x)$, from which the predictive distribution $p_\theta(y \mid x)$ is obtained via the softmax function.
The model's predictive entropy for a sample $x$ is defined as:
\begin{equation}
H\!\left(p_\theta(\cdot \mid x)\right) = - \sum_{k=1}^{K} p_\theta(y = k \mid x) \log p_\theta(y = k \mid x).
\end{equation}

\noindent\textbf{Tent}~\citep{wang2021tent} performs test-time adaptation by minimizing prediction entropy on an unlabeled target batch $B_t$:
\begin{equation}
\mathcal{L}_{\mathrm{Tent}} = \frac{1}{|B_t|} \sum_{x \in B_t} H\!\left(p_\theta(\cdot \mid x)\right).
\end{equation}
By minimizing predictive entropy, Tent encourages confident predictions on target samples under distribution shift. At test time, adaptation is restricted to the affine parameters of normalization layers, while the remaining network parameters are held fixed.\\

\noindent\textbf{SHOT}~\citep{liang2020we} assumes source-free adaptation and keeps the classifier $h_w$ fixed while adapting only the target feature extractor $g_\phi$. Its objective consists of three terms:
\begin{equation}
\mathcal{L}_{\mathrm{SHOT}} = \mathcal{L}_{\mathrm{ent}} + \mathcal{L}_{\mathrm{div}} + \beta \mathcal{L}_{\mathrm{PL}},
\end{equation}
where,
\begin{equation}
\mathcal{L}_{\mathrm{ent}}
=
-
\mathbb{E}_{x \sim \mathcal{X}_t}
\sum_{k=1}^{K}
p_k(x)\log p_k(x),
\end{equation}
\begin{equation}
\mathcal{L}_{\mathrm{div}}
=
\sum_{k=1}^{K} \hat{p}_k \log \hat{p}_k,
\qquad
\hat{p}_k = \mathbb{E}_{x \sim \mathcal{X}_t}[p_k(x)].
\end{equation}
Here, $\mathcal{L}_{\mathrm{ent}}$ encourages confident predictions on target samples, $\mathcal{L}_{\mathrm{div}}$ promotes diversity in the marginal output distribution and prevents degenerate collapse to a single class, and $\mathcal{L}_{\mathrm{PL}}$ denotes the pseudo-label loss.\\

\noindent\textbf{T3A}~\citep{iwasawa2021testtime} is an optimization-free method that keeps the feature extractor $g_\phi$ fixed and adapts the classifier online using target features. Formally, let $z = g_\phi(x)$ denote the feature representation of $x$, and let $\{\omega_k\}_{k=1}^K$ denote the source classifier weight vectors. For each class $k$, T3A maintains a support set $S_k^{(t)}$ at test-time step $t$, consisting of target feature vectors assigned to that class.
The class template is then computed as the mean of the corresponding support set:

\begin{equation}
c_k^{(t)} = \frac{1}{|S_k^{(t)}|} \sum_{z \in S_k^{(t)}} z.
\end{equation}

\noindent Prediction is then performed using the adjusted classifier
\begin{equation}
p(y = k \mid z) \propto \exp\!\left(z^\top c_k^{(t)}\right),
\end{equation}

where $c_k^{(t)}$ is the class prototype for class $k$ at time $t$. In this way, T3A adapts the classifier geometry directly by refining class prototypes from target test features, without updating network parameters through gradient-based optimization.

\subsection{NeuroAdapt-Bench Design}
\label{sec:bench_design}
   \begin{figure}[t]
     \centering
     \includegraphics[width=\textwidth]{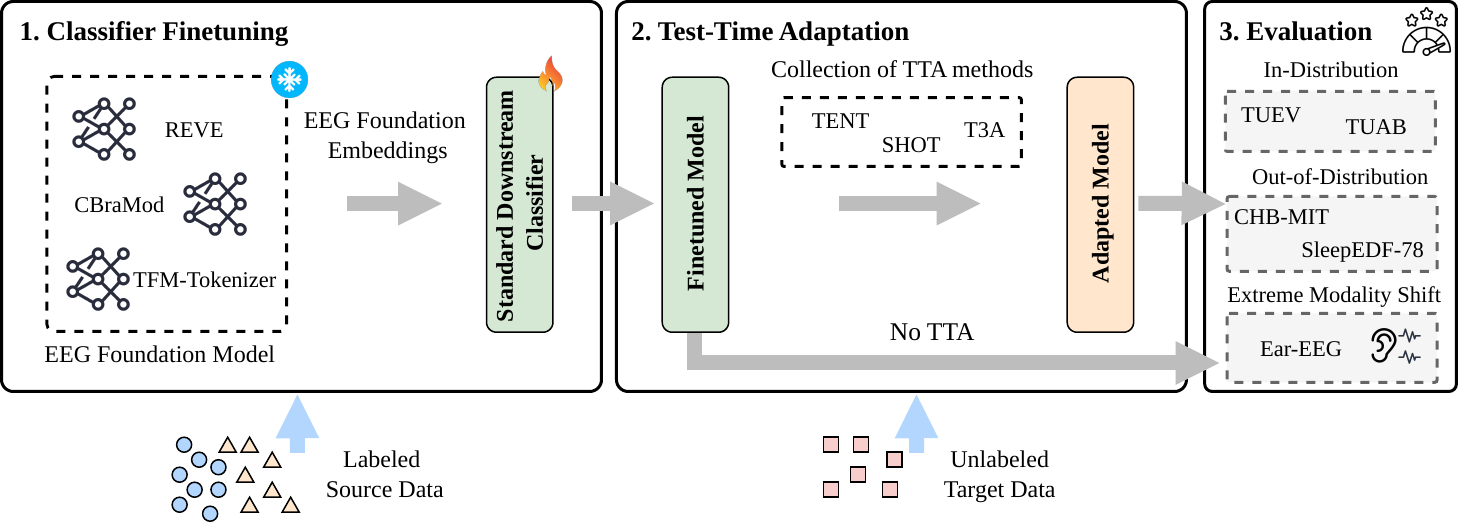}

     \caption{Overview of NeuroAdapt-Bench.
     The benchmark consists of three stages: (1) supervised finetuning of an EEG foundation model (e.g., REVE, CBRaMod, or TFM-Tokenizer) with a classification head on labeled source-domain data; (2) optional test-time adaptation on unlabeled target-domain data using methods such as TENT, SHOT, or T3A, alongside a no-adaptation baseline; and (3) evaluation on target-domain samples to measure performance and robustness under distribution shift.}
     \label{fig:method_figure}
   \end{figure}

As illustrated in Figure~\ref{fig:method_figure}, NeuroAdapt-Bench follows a three-stage pipeline: (1) classifier fine-tuning, (2) test-time adaptation, and (3) evaluation.

\paragraph{Stage 1: Classifier Fine-Tuning}

Each foundation model is paired with the same lightweight classification head, replacing any model-specific heads used in prior work, to control for classifier architecture as a confounding factor in cross-model comparison. This design ensures that downstream performance differences are more cleanly attributable to the pretrained encoder representations rather than to gains induced by model-specific classification layers. The encoder backbone is frozen, and only the classification head is trained, further standardizing the optimization setting across models and preserving a consistent initialization for subsequent test-time adaptation. Architectural details of the shared classifier head are provided in Appendix~\ref{sec:shared Shared Downstream Classifier and Fine-Tuning}. Model selection is performed exclusively on a held-out validation split, with no access to test data.

\paragraph{Stage 2: Test-Time Adaptation}
The held-out test split is treated as unlabeled target data. During adaptation, only EEG signals are provided to the model, and ground-truth labels are strictly withheld. As mentioned previously, we evaluate three representative TTA methods alongside a No-TTA baseline, selected to span two orthogonal axes of clinical deployment constraints: whether the full target set must be available before adaptation begins (offline vs.\ online), and whether the method requires gradient computation (Table~\ref{tab:tta-methods}).

\noindent\textbf{No-TTA} performs inference with the frozen fine-tuned checkpoint and serves as the unadapted baseline. \textbf{Tent} updates the affine parameters of normalization layers (batch normalization, layer normalization, and group normalization) through entropy minimization on each incoming batch, accumulating state across the test stream without requiring a prior pass over the full target set. \textbf{T3A} maintains a per-class support set of low-entropy prototype features that is updated incrementally with each batch, and no gradient computation is required. \textbf{SHOT} first performs a full pass over the target set to construct refined feature centroids via mutual information maximization and pseudo-labeling, and then adapts the encoder through gradient descent. It therefore requires the complete target set to be available before adaptation begins.

Online methods (Tent, T3A) are suitable for streaming scenarios such as continuous bedside monitoring, whereas offline methods (SHOT) is better suited to settings in which a batch of recordings can be collected before deployment (e.g. sleep studies).

\paragraph{Stage 3: Evaluation}

After adaptation, ground-truth labels are used to compute standard classification metrics, including accuracy, balanced accuracy, ROC-AUC, PR-AUC, Cohen’s $\kappa$, and weighted $F_1$. For each (method, model, dataset) combination, we report the mean and standard deviation across five random seeds. To isolate the effect of adaptation, we additionally report the relative improvement:
\begin{equation}
\Delta_{\text{TTA}} = \text{metric}_{\text{TTA}} - \text{metric}_{\text{No-TTA}},
\end{equation}
computed per seed prior to aggregation. This ensures that the reported variability reflects differences in adaptation performance rather than absolute model accuracy. Appendix~\ref{sec:appendix_percentage} reports the same comparisons as relative changes, and Appendix~\ref{sec_appendix_significance} reports statistical significance tests against No-TTA. Appendix~\ref{sec:appendix_compute_cost} reports measured adaptation runtime and peak GPU memory.

\subsection{Experiment Setup}
\label{sec:eval_setup}

We evaluate TTA for EEG foundation models using a standardized downstream pipeline that isolates the effect of deployment-time adaptation from model-specific classifier design. The benchmark spans four foundation-model variants, five EEG datasets, both binary and multiclass tasks, and patient-disjoint evaluation.

\paragraph{Datasets, Tasks and Metrics.}
We evaluate on five EEG datasets: TUEV~\citep{harati2015improved}, TUAB~\citep{lopez2015automated}, CHB-MIT~\citep{Shoeb2010ApplicationOM} from PhysioNet~\citep{goldberger2000physiobank}, EarEEG~\citep{bjarke2025ear,ds005178:1.0.0}, and SleepEDF-78~\citep{N9/EUHGHS_2022}. TUEV and TUAB are treated as in-distribution datasets, as they are included in the pretraining corpora of the evaluated EEG foundation models and correspond to event classification and abnormality detection tasks. In contrast, CHB-MIT and SleepEDF-78 are considered out-of-distribution because they are not part of the pretraining data for most models, except for TFM-Tokenizer, which includes CHB-MIT during pretraining. CHB-MIT focuses on epilepsy seizure detection, and SleepEDF-78 focuses on sleep staging. To further study robustness to extreme distributional shift, we include an ear-EEG setting that differs substantially in signal modality, channel configuration, and acquisition setup from those seen during pretraining for all models. This setting focuses on sleep staging using EarEEG data and represents a challenging out-of-domain evaluation scenario.

We analyze these settings as subject-level, cross-dataset/task, and cross-modality/device shifts rather than as one uniform distribution shift. All evaluations use patient-disjoint splits to avoid subject leakage. To illustrate the evaluated distribution shifts, Appendix~\ref{sec:appendix_visualization} provides a qualitative t-SNE diagnostic of source-target structure in raw EEG and frozen foundation-model embeddings before TTA.

We report balanced accuracy, ROC-AUC, and PR-AUC for binary classification tasks, and balanced accuracy, Cohen's $\kappa$, and weighted $F_1$ for multi-class classification.

\paragraph{EEG Foundation Models.}
We evaluate three EEG foundation-model families, including CBraMod~\citep{wang2024cbramod}, TFM-Tokenizer~\citep{pradeepkumar2026tokenizing}, and REVE~\citep{ouahidi2025reve}. For REVE, we consider both the Base and Large variants, as it is pretrained on one of the largest EEG corpora to date. CBraMod provides an efficient architecture with demonstrated generalization across multiple EEG tasks. In contrast to these continuous embedding-based models, TFM-Tokenizer introduces a discrete tokenization framework for EEG, enabling evaluation across fundamentally different representation paradigms. All models are integrated through a shared interface while preserving their native input processing and representational assumptions. We attach a common lightweight classification head to the publicly available and fixed pretrained backbones and fine-tune one classifier per (model, dataset) pair. Test-time adaptation is then applied during inference.

\paragraph{Standard Downstream Classifier.}
For fair comparison, each pretrained backbone is paired with the same lightweight downstream classifier. The encoder produces a latent representation, which is pooled when needed, passed through a shared feature adapter, and mapped to task logits by a linear classification layer. This removes backbone-specific downstream engineering as a major confound.

\section{Results and Discussion}
\label{sec:results}

\subsection{Does test-time adaptation improve performance for EEG foundation models?}

\begin{figure}[t]
    \centering
    \includegraphics[width=\linewidth]{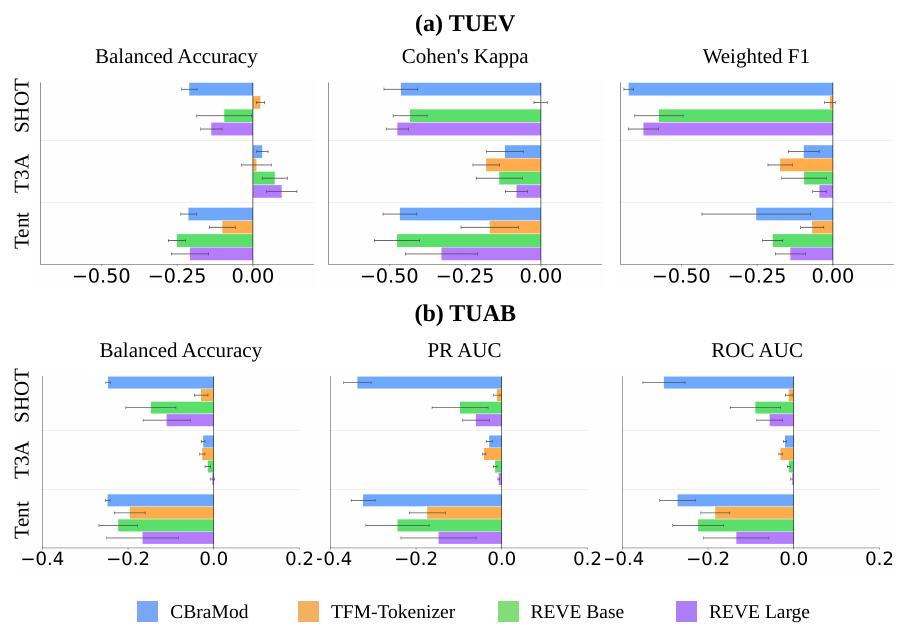}
    \caption{TTA relative performance on in-distribution datasets (TUEV and TUAB). (a) $\Delta_{\text{TTA}}$ on TUEV relative to the No-TTA baseline; (b) $\Delta_{\text{TTA}}$ on TUAB relative to the No-TTA baseline.}
    \label{fig:in_dist_delta}
\end{figure}

Figure~\ref{fig:in_dist_delta} shows the relative performance of TTA methods compared to the No-TTA baseline on TUEV and TUAB, which are in-distribution datasets included in the pretraining corpora of the EEG foundation models. In this setting, the primary source of variability is subject-level differences between the training and test splits. Across models and datasets, the evaluated gradient-based methods (Tent and SHOT) consistently degrade performance, often substantially. In contrast, T3A exhibits the most stable behavior and provides modest improvements in balanced accuracy on TUEV, with lower variability across seeds and batch sizes. On TUAB, however, all evaluated TTA methods degrade performance, with Tent showing the largest drop.

These results show limited benefit from the evaluated TTA methods on the in-distribution datasets. We avoid attributing this pattern to a specific mechanism, since we do not directly measure representation alignment or adaptation-induced drift in this experiment. The relative stability of T3A is therefore treated as an empirical observation rather than evidence that its optimization-free design is the causal factor.

\subsection{How does TTA behave under cross-dataset and task shifts?}

\begin{figure}[t]
    \centering
    \includegraphics[width=\linewidth]{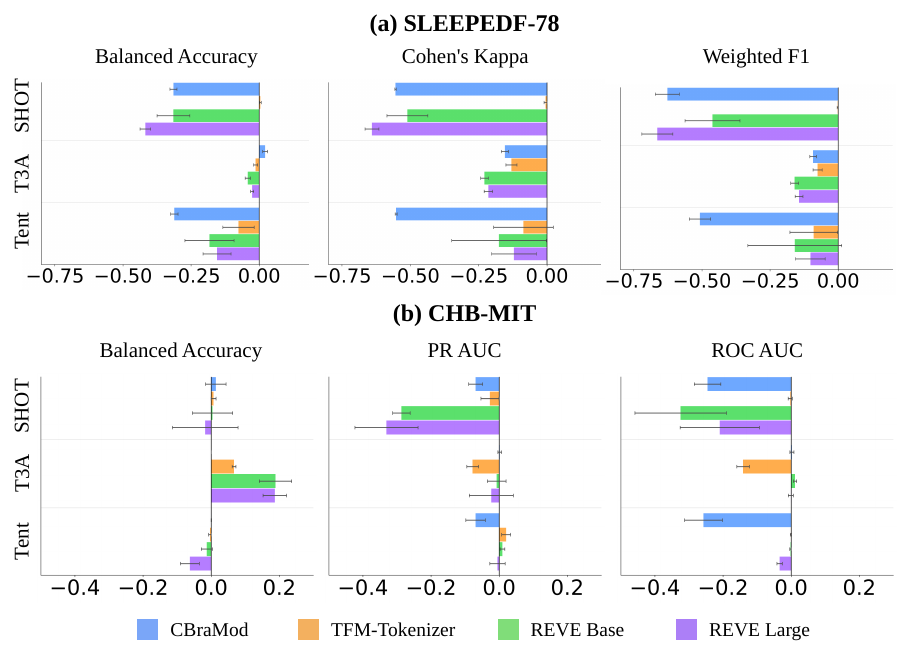}
    \caption{TTA relative performance on out-of-distribution datasets (SLEEPEDF-78 and CHB-MIT). (a) $\Delta_{\text{TTA}}$ on SLEEPEDF-78 relative to the No-TTA baseline; (b) $\Delta_{\text{TTA}}$ on CHB-MIT relative to the No-TTA baseline.}
    \label{fig:out_of_dist_delta}
\end{figure}

To evaluate TTA under realistic distribution shifts, we consider out-of-distribution datasets that are not included in the pretraining corpora of most evaluated models. These settings introduce compounded shifts that vary simultaneously across cohort, acquisition protocol, hardware, and sensor configuration. We therefore characterize robustness under these combined shifts rather than attempting to isolate a single dominant cause, which would require controlled, metadata-specific analyses beyond the scope of this benchmark.

Figure~\ref{fig:out_of_dist_delta} summarizes the relative performance of TTA methods compared to the No-TTA baseline on SleepEDF-78 and CHB-MIT. On CHB-MIT, T3A provides consistent improvements in balanced accuracy across most models, showing trends similar to those observed in the in-distribution setting. We do not test the mechanism underlying this pattern directly. We therefore treat the T3A result as an empirical trend rather than evidence of a specific causal mechanism.

The REVE family, in particular, benefits from T3A on balanced accuracy and shows less to no degradation in ROC-AUC and PR-AUC. However, REVE shows higher degradation in PR-AUC and ROC AUC under SHOT.

In contrast, TTA on the more challenging SleepEDF-78 dataset shows greater degradation across nearly all methods and metrics. This dataset differs substantially in task (sleep staging) and channel configuration, making adaptation particularly difficult. T3A offers only marginal gains (e.g., slight improvements in balanced accuracy for CBraMod), while Tent and SHOT consistently degrade performance. Notably, TFM-Tokenizer shows relatively greater robustness across all TTA approaches, with smaller performance drops compared to other models. Overall, these results demonstrate that the evaluated TTA methods struggle to generalize under cross-dataset shifts.

\subsection{Can TTA handle unseen EEG modalities such as EarEEG?}

In this section, we evaluate TTA under extreme distribution shift by considering an unseen EEG modality, such as ear-EEG. Most EEG foundation models are pretrained on scalp EEG data following the standard 10-20 system, whereas ear-EEG differs substantially in signal characteristics, channel configuration, and acquisition setup. With the growing adoption of wearable EEG technologies~\citep{bjarke2025ear}, understanding cross-modality generalization from scalp-EEG to wearable EEG has become increasingly important~\citep{anandakumar2023knowledge}.

In this setting, we assess whether the evaluated TTA methods can adapt pretrained scalp-EEG foundation models to ear-EEG sleep staging and the relative improvement results are summarized in Table~\ref{tab:eareeg_delta}. Overall, the evaluated TTA methods are unstable under this modality shift. The evaluated gradient-based approaches (SHOT and Tent) consistently degrade performance across models and metrics. In contrast, T3A is more stable and yields improvements for some models, notably for CBraMod across all metrics, with moderate gains for REVE in balanced accuracy.

\begin{table}[t]
  \centering
  \caption{TTA performance on the Ear-EEG sleep staging task (EESM23). We report $\Delta_{\text{TTA}}$ relative to the No-TTA baseline for each foundation model, aggregated across random seeds and adaptation batch sizes. Values are shown as mean $\pm$ standard deviation.}

  \begin{tabular}{llccc}
    \toprule
    \textbf{\shortstack{TTA \\ Method}} & \textbf{\shortstack{Foundation\\Model}} & \textbf{\shortstack{Balanced\\Acc.$\Delta$}} & \textbf{\shortstack{Cohen's\\Kappa$\Delta$}} & \textbf{\shortstack{Weighted\\F1$\Delta$}} \\
    \midrule
    \multirow{4}{*}{SHOT} & CBraMod & $-0.065 \pm 0.020$ & $-0.087 \pm 0.027$ & $-0.224 \pm 0.015$ \\
     & TFM-Tokenizer & $+0.000 \pm 0.000$ & $-0.000 \pm 0.001$ & $-0.000 \pm 0.000$ \\
     & REVE-Base & $-0.018 \pm 0.032$ & $-0.085 \pm 0.064$ & $-0.119 \pm 0.076$ \\
     & REVE-Large & $-0.068 \pm 0.056$ & $-0.156 \pm 0.083$ & $-0.150 \pm 0.087$ \\
    \midrule
    \multirow{4}{*}{T3A} & CBraMod & $\textbf{+0.048} \pm 0.012$ & $\textbf{+0.064} \pm 0.016$ & $\textbf{+0.018} \pm 0.009$ \\
     & TFM-Tokenizer & $-0.005 \pm 0.007$ & $-0.042 \pm 0.006$ & $-0.009 \pm 0.006$ \\
     & REVE-Base & $+0.037 \pm 0.007$ & $+0.001 \pm 0.015$ & $+0.001 \pm 0.017$ \\
     & REVE-Large & $+0.022 \pm 0.007$ & $-0.010 \pm 0.014$ & $-0.009 \pm 0.010$ \\
    \midrule
    \multirow{4}{*}{Tent} & CBraMod & $-0.064 \pm 0.022$ & $-0.084 \pm 0.030$ & $-0.189 \pm 0.060$ \\
     & TFM-Tokenizer & $-0.001 \pm 0.001$ & $+0.000 \pm 0.001$ & $-0.000 \pm 0.001$ \\
     & REVE-Base & $-0.032 \pm 0.022$ & $-0.047 \pm 0.035$ & $-0.046 \pm 0.028$ \\
     & REVE-Large & $-0.018 \pm 0.014$ & $-0.025 \pm 0.018$ & $-0.019 \pm 0.015$ \\
    \bottomrule
  \end{tabular}
  \label{tab:eareeg_delta}
\end{table}

\subsection{How does adaptation batch size affect performance?}

\begin{figure}[b]
    \centering
    \includegraphics[width=\linewidth]{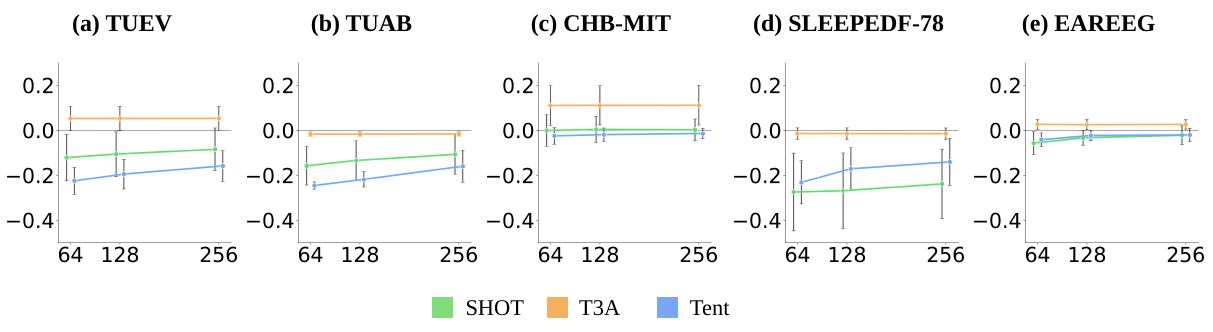}
    \caption{Balanced accuracy improvements relative to the No-TTA baseline across adaptation batch sizes (64, 128, 256). Results are shown for (a) TUEV, (b) TUAB, (c) CHB-MIT, (d) SleepEDF-78, and (e) EarEEG.}
    \label{fig:batch_size_ablation}
\end{figure}

Figure~\ref{fig:batch_size_ablation} shows the effect of adaptation batch size on balanced accuracy across all models and datasets. Overall, increasing batch size for the evaluated methods does not provide consistent performance gains. For the evaluated gradient-based methods (SHOT and Tent), increasing batch size can reduce degradation in some settings, but it does not provide consistent gains across datasets.

In contrast, T3A is relatively insensitive to batch size, as it updates class prototypes without gradient-based optimization. These results suggest that simply increasing batch size is insufficient to stabilize or improve the evaluated TTA methods for EEG.

\subsection{Which TTA methods are most stable?}
Across the evaluated TTA methods, T3A was by far the most stable, as when used with foundation models, it caused the least degradation relative to the No-TTA baseline and occasionally even yielded benefits. We hypothesize that, as an optimization-free approach, T3A avoids updating model parameters, which may help preserve pretrained representations under heterogeneous EEG deployment settings. However, more controlled experiments across other approaches would be necessary to validate this mechanism.

In contrast, the evaluated gradient-based methods (SHOT and Tent) frequently lead to substantial performance degradation, often outweighing their occasional benefits on specific model architectures. We hypothesize that their objectives may perturb pretrained representations, which would be consistent with the negative transfer observed under distribution shift, although we similarly do not verify this mechanism directly. Validating the dominant mechanism is left to future work.

Overall, our ablation suggests that increasing batch size generally helps reduce degradation for the evaluated TTA methods. However, the performance benefits do not seem large enough to justify increasing memory or computational requirements without a significant gain.

\paragraph{How do online and offline adaptation strategies compare?}

In terms of the online versus offline characteristics outlined in Table~\ref{tab:tta-methods}, we find that, for the evaluated TTA methods, the distinction between online and offline adaptation plays a less significant role than the nature of the update mechanism. While both Tent and T3A operate in an online setting, their behaviors differ. Tent updates internal normalization parameters, often introducing instability, whereas T3A modifies class prototypes without updating model weights, resulting in more stable performance and occasional improvements. These results suggest that the where and how of adaptation are more critical for these methods than whether it is performed in a streaming or batch setting.

\paragraph{Does representation type affect TTA behavior?}

Overall, our experiments on the evaluated models and methods suggest responses to adaptation vary between continuous-embedding and discrete-tokenization models. However, we present this as a hypothesis suggested by the empirical trends, and leave a controlled test to future work.

For instance, TFM-Tokenizer shows more resistance to degradation for the evaluated TTA methods, particularly under the SHOT method, across all datasets in both in-distribution and out-of-distribution settings. Other continuous embedding-based models, such as REVE, seem to benefit the most from the T3A method, especially on the CHB-MIT dataset.

\subsection{Discussion and Implications for Future Works}

Our results highlight several important considerations for deploying TTA with EEG foundation models. First, the evaluated foundation models are not plug-and-play in real-world clinical settings, while they perform well on in-distribution datasets (e.g., TUAB, TUEV), performance degrades substantially under out-of-distribution conditions, particularly under extreme shifts such as EarEEG. This underscores the distribution shift as a primary barrier to reliable deployment. Second, we observe that the evaluated optimization-free TTA methods exhibit greater stability than the evaluated gradient-based approaches, which exhibit performance degradation. This suggests that future work should prioritize robustness and explore alternative adaptation strategies that minimize disruptive updates.

\subsection{Limitations}

Our benchmark covers three representative test-time adaptation methods across batch and streaming settings, four EEG foundation model variants, and five downstream datasets. However, it does not span the full space of TTA approaches and model families, so our conclusions are scoped to the evaluated EEG foundation models and TTA categories. We benchmark generic TTA methods rather than physiology-aware adaptation methods designed for EEG-specific nonstationarity, subject variability, spectral structure, and low signal-to-noise ratio.

Our work provides a reproducible evaluation framework to test and study the effectiveness of TTA methods with EEG foundation models. We do not evaluate calibration or uncertainty, so reliability diagrams, expected calibration error, and uncertainty-aware evaluation remain important before adapted models can be considered clinically trustworthy.

Although our settings include patient-disjoint, cross-dataset, cross-task/institutional, and cross-modality shifts, stronger real-world shifts such as temporal drift and center-specific acquisition protocols remain future stress tests.

We also restrict this study to EEG foundation models, and do not include analysis of non-foundation or general-purpose time-series foundation model baselines which require broader experimental controls. Such broader analyses could provide the basis for future work, but are not considered in this benchmark.

Finally, computational cost is another limitation. Larger models, such as REVE-Large, can make adaptation memory-intensive, limiting the feasible batch sizes and hardware accessibility. Evaluating computational efficiency and the feasibility of clinical deployment more directly is an important direction for future work.

\section{Conclusion}
In this work, we present \emph{NeuroAdapt-Bench}, a systematic benchmark for evaluating test-time adaptation methods with EEG foundation models under realistic distribution shifts. Across diverse datasets and tasks, we find that the evaluated TTA methods yield inconsistent gains and often degrade performance relative to the No-TTA baseline. In particular, the evaluated optimization-free approaches such as T3A demonstrate greater stability and often positive gains, whereas the evaluated gradient-based methods are more prone to degradation.

Our results highlight that distribution shift remains a fundamental challenge for deploying the evaluated EEG foundation models, and that the evaluated TTA methods from other domains do not transfer reliably in this benchmark, thereby motivating EEG-specific test-time adaptation approaches. We frame these results as diagnostic, not as a final rejection of TTA for EEG. Overall, our study establishes a reproducible evaluation framework and provides empirical insights to guide the development of reliable and robust adaptation strategies for EEG.

\bibliography{main}

\newpage
\appendix
\section{}

\subsection{Dataset Preprocessing and Split Policy}
\label{sec:Dataset_Preprocessing}

Table~\ref{tab:appendix_dataset_preprocessing} summarizes the dataset-specific preprocessing and split policy used in the benchmark. Across all datasets, preprocessing standardizes temporal support and amplitude scaling while preserving dataset-specific channel geometry rather than forcing all recordings into a single global montage. For each foundation model, we reproduce the preprocessing expected by the original model implementation, including its resampling, windowing, normalization, and channel-handling conventions.

  \begin{table}[H]
    \centering
    \caption{Dataset preprocessing and split policy used in the benchmark.}
    \label{tab:appendix_dataset_preprocessing}
    \scriptsize
    \setlength{\tabcolsep}{3pt}
    \begin{adjustbox}{max width=\textwidth}
    \begin{tabular}{p{1.8cm}p{1.3cm}p{1.0cm}p{1.3cm}p{1.2cm}p{0.9cm}p{2.5cm}p{5.0cm}}
    \toprule
    \textbf{Dataset} & \textbf{Task} & \textbf{Classes} & \textbf{Channels} & \textbf{Window} & \textbf{Rate} & \textbf{Normalization} &
  \textbf{Split policy} \\
    \midrule
    TUEV & Multiclass & 6 & 16 & 5 s & 200 Hz & Per-channel 95th percentile & Official train/eval pools; seeded patient-level 80/20 split
  of the train pool into train/val \\
    TUAB & Binary & 2 & 16 & 10 s & 200 Hz & Per-channel 95th percentile & Official train/eval pools; seeded patient-level 80/20 split of
  the train pool into train/val \\
    CHB-MIT & Binary & 2 & 16 & 10 s & 256 Hz & Per-channel 95th percentile & Precomputed train/val/test split \\
    SleepEDF-78 & Multiclass & 5 & 2 & 30 s & 100 Hz & Per-channel 95th percentile & Precomputed train/val/test split \\
    EarEEG & Multiclass & 6 & 4 & 30 s & 250 Hz & Per-channel 95th percentile & Precomputed train/val/test split; final stored channel
  dropped before preprocessing \\
    \bottomrule
    \end{tabular}
    \end{adjustbox}
  \end{table}

Table~\ref{tab:dataset_task_summary} summarizes the benchmark datasets, tasks, and shift categories. We separate these categories to avoid treating subject-level variation, cross-dataset/task transfer, and device/modality transfer as one uniform shift.

  \begin{table}[H]
    \centering
    \caption{Dataset-task summary and shift category.}
    \label{tab:dataset_task_summary}
    \scriptsize
    \setlength{\tabcolsep}{5pt}
    \renewcommand{\arraystretch}{1.1}
    \begin{adjustbox}{max width=\textwidth}
    \begin{tabular}{llll}
    \toprule
    \textbf{Dataset} & \textbf{Task} & \textbf{Classes} & \textbf{Shift category} \\
    \midrule
    TUEV & Event classification & 6 & Subject-level (in-distribution) \\
    TUAB & Abnormality detection & 2 & Subject-level (in-distribution) \\
    CHB-MIT & Seizure detection & 2 & Cross-dataset/task (out-of-distribution) \\
    SleepEDF-78 & Sleep staging & 5 & Cross-dataset/task (out-of-distribution) \\
    EarEEG & Sleep staging & 6 & Cross-modality/device (extreme modality shift)\\
    \bottomrule
    \end{tabular}
    \end{adjustbox}
  \end{table}

\subsection{Shared Downstream Classifier and Fine-Tuning}
\label{sec:shared Shared Downstream Classifier and Fine-Tuning}

To compare pretrained backbones under a common downstream protocol, each foundation model is paired with the same lightweight classifier head. The shared head first applies LayerNorm to the pooled feature representation, then projects features to a 128-dimensional hidden space, applies GELU and dropout, and finally maps to task logits with a linear classification layer. Encoder backbones are frozen during the reported downstream fine-tuning experiments. REVE is the only backbone that additionally consumes channel-position information, while the other backbones operate only on the waveform input. For REVE, channel positions are constructed from a shared 3D electrode position bank; for bipolar channels, the benchmark uses the mean of the two endpoint electrode coordinates. For EarEEG, the four channels are mapped to the aliases A2, T8, A1, and T7 before position lookup. CBraMod additionally applies a fixed input scaling factor of 100 to match the expected amplitude range of the pretrained backbone.
  \begin{table}[H]
    \centering
    \caption{Shared downstream classifier and fine-tuning configuration.}
    \label{tab:appendix_finetuning}
    \scriptsize
    \setlength{\tabcolsep}{4pt}
    \begin{adjustbox}{max width=\textwidth}
    \begin{tabular}{p{3.6cm}p{9.9cm}}
    \toprule
    \textbf{Component} & \textbf{Setting} \\
    \midrule
    Foundation-model variants & CBraMod, TFM, REVE-Base, REVE-Large \\
    Encoder training during fine-tuning & Frozen \\
    Shared downstream head & LayerNorm $\rightarrow$ Linear(128) $\rightarrow$ GELU $\rightarrow$ Dropout(0.1) $\rightarrow$ Linear($C$) \\
    Pooling policy & Mean pooling for sequence outputs; native pooled outputs used when provided by the backbone \\
    Loss & Cross-entropy objective \\
    Optimizer & AdamW \\
    Learning rate & $10^{-3}$ \\
    Weight decay & $10^{-4}$ \\
    Epochs & 10 \\
    Classifier training batch size & 512 \\
    Data augmentation & None \\
    Model selection for binary tasks & Best validation ROC-AUC \\
    Model selection for multiclass tasks & Best validation Cohen's $\kappa$ \\
    Reported study seeds & Five random seeds used in the reported study \\
    \bottomrule
    \end{tabular}
    \end{adjustbox}
  \end{table}

\subsection{Test-Time Adaptation Configuration}

Table~\ref{tab:appendix_tta_config} summarizes the operational configuration of the evaluated test-time adaptation methods. We report the specific settings used in the benchmark rather than the full space of possible variants for each method family. We do not tune TTA hyperparameters per dataset because target labels are withheld and per-dataset tuning would leak target information. Instead, we use fixed settings based on the original method defaults and vary only adaptation batch size systematically in the batch-size ablation.

  \begin{table}[H]
    \centering
    \caption{Test-time adaptation configuration used in the benchmark.}
    \label{tab:appendix_tta_config}
    \scriptsize
    \setlength{\tabcolsep}{3pt}
    \begin{adjustbox}{max width=\textwidth}
    \begin{tabular}{p{1.2cm}p{1.5cm}p{3.7cm}p{1.5cm}p{5.6cm}}
    \toprule
    \textbf{Method} & \textbf{Regime} & \textbf{Updated component} & \textbf{Optimizer} & \textbf{Key settings} \\
    \midrule
    No-TTA & None & None & None & Frozen checkpoint inference \\
    Tent & Online & Affine parameters of normalization layers & SGD & lr $=10^{-3}$, momentum $=0.9$, steps $=1$, episodic $=$ False \\
    SHOT & Offline & Trainable feature modules only; classifier fixed & SGD & lr $=10^{-4}$, wd $=10^{-4}$, steps $=1$, episodic $=$
  False, MI weight $=1.0$, PL weight $=1.0$ \\
    T3A & Online & Classifier supports / prototypes & None & filter\_k $=20$, episodic $=$ False \\
    \bottomrule
    \end{tabular}
    \end{adjustbox}
  \end{table}

\subsection{Evaluation and Aggregation}

Table~\ref{tab:appendix_eval_config} summarizes the evaluation settings used throughout the benchmark.

  \begin{table}[H]
    \centering
    \caption{Evaluation and aggregation settings.}
    \label{tab:appendix_eval_config}
    \scriptsize
    \setlength{\tabcolsep}{4pt}
    \begin{adjustbox}{max width=\textwidth}
    \begin{tabular}{p{3.8cm}p{9.5cm}}
    \toprule
    \textbf{Setting} & \textbf{Value} \\
    \midrule
    Primary binary metrics & Balanced accuracy, ROC-AUC, PR-AUC \\
    Primary multiclass metrics & Balanced accuracy, Cohen's $\kappa$, weighted $F_1$ \\
    Additional logged metric & Accuracy \\
    Aggregation & Mean $\pm$ standard deviation over study seeds \\
    Relative adaptation metric & $\Delta_{\mathrm{TTA}} = \mathrm{metric}_{\mathrm{TTA}} - \mathrm{metric}_{\mathrm{No\text{-}TTA}}$,
  computed per seed before averaging \\
    TTA evaluation batch sizes & 64, 128, 256 \\
    \bottomrule
    \end{tabular}
    \end{adjustbox}
  \end{table}

\subsection{Per-Dataset Delta Tables}
\label{sec:appendix_delta_tables}

  Our quantitative results are summarized in the following tables. Each table reports
  performance deltas relative to the No-TTA baseline for a single dataset, separated by
  foundation model and aggregated across seeds and adaptation batch sizes. Values are
  mean $\pm$ standard deviation, and bold values indicate the largest mean for each
  metric in each table.

\begin{table}[H]
  \centering
  \scriptsize
  \setlength{\tabcolsep}{3pt}
  \renewcommand{\arraystretch}{1.05}
  \caption{TUEV delta relative to the no-TTA baseline.}
  \label{tab:tuev_delta}
  \begin{tabular}{p{1.45cm}p{1.75cm}ccc}
    \toprule
    \textbf{\shortstack{TTA\\Method}} & \textbf{\shortstack{Foundation\\Model}} & \textbf{Bal. Acc. $\Delta$} & \textbf{Cohen's $\kappa$ $\Delta$} & \textbf{Weighted F1 $\Delta$} \\
    \midrule
    \multirow{4}{*}{\vspace{-10pt}SHOT} & CBraMod & $-0.210 \pm 0.025$ & $-0.462 \pm 0.056$ & $-0.673 \pm 0.016$ \\
    \cmidrule(lr){2-5}
     & TFM & $+0.024 \pm 0.014$ & $\mathbf{-0.001} \pm 0.022$ & $\mathbf{-0.010} \pm 0.018$ \\
    \cmidrule(lr){2-5}
     & REVE-Base & $-0.095 \pm 0.090$ & $-0.431 \pm 0.056$ & $-0.574 \pm 0.080$ \\
     & REVE-Large & $-0.137 \pm 0.035$ & $-0.473 \pm 0.037$ & $-0.624 \pm 0.050$ \\
    \cmidrule(lr){1-5}
    \multirow{4}{*}{\vspace{-10pt}T3A} & CBraMod & $+0.031 \pm 0.019$ & $-0.118 \pm 0.061$ & $-0.096 \pm 0.051$ \\
    \cmidrule(lr){2-5}
     & TFM & $+0.012 \pm 0.049$ & $-0.181 \pm 0.044$ & $-0.174 \pm 0.040$ \\
    \cmidrule(lr){2-5}
     & REVE-Base & $+0.072 \pm 0.042$ & $-0.137 \pm 0.076$ & $-0.095 \pm 0.074$ \\
     & REVE-Large & $\mathbf{+0.095} \pm 0.050$ & $-0.081 \pm 0.037$ & $-0.044 \pm 0.022$ \\
    \cmidrule(lr){1-5}
    \multirow{4}{*}{\vspace{-10pt}Tent} & CBraMod & $-0.212 \pm 0.026$ & $-0.465 \pm 0.056$ & $-0.253 \pm 0.179$ \\
    \cmidrule(lr){2-5}
     & TFM & $-0.101 \pm 0.043$ & $-0.169 \pm 0.095$ & $-0.068 \pm 0.038$ \\
    \cmidrule(lr){2-5}
     & REVE-Base & $-0.250 \pm 0.029$ & $-0.474 \pm 0.075$ & $-0.198 \pm 0.033$ \\
     & REVE-Large & $-0.208 \pm 0.061$ & $-0.328 \pm 0.119$ & $-0.140 \pm 0.050$ \\
    \bottomrule
  \end{tabular}
\end{table}

\begin{table}[H]
  \centering
  \scriptsize
  \setlength{\tabcolsep}{3pt}
  \renewcommand{\arraystretch}{1.05}
  \caption{TUAB delta relative to the no-TTA baseline.}
  \label{tab:tuab_delta}
  \begin{tabular}{p{1.45cm}p{1.75cm}ccc}
    \toprule
    \textbf{\shortstack{TTA\\Method}} & \textbf{\shortstack{Foundation\\Model}} & \textbf{Bal. Acc. $\Delta$} & \textbf{ROC AUC $\Delta$} & \textbf{PR AUC $\Delta$} \\
    \midrule
    \multirow{4}{*}{\vspace{-10pt}SHOT} & CBraMod & $-0.247 \pm 0.005$ & $-0.303 \pm 0.049$ & $-0.337 \pm 0.032$ \\
    \cmidrule(lr){2-5}
     & TFM & $-0.030 \pm 0.016$ & $-0.012 \pm 0.008$ & $-0.011 \pm 0.007$ \\
    \cmidrule(lr){2-5}
     & REVE-Base & $-0.147 \pm 0.058$ & $-0.090 \pm 0.059$ & $-0.097 \pm 0.065$ \\
     & REVE-Large & $-0.110 \pm 0.055$ & $-0.056 \pm 0.030$ & $-0.060 \pm 0.032$ \\
    \cmidrule(lr){1-5}
    \multirow{4}{*}{\vspace{-10pt}T3A} & CBraMod & $-0.025 \pm 0.004$ & $-0.021 \pm 0.004$ & $-0.029 \pm 0.007$ \\
    \cmidrule(lr){2-5}
     & TFM & $-0.027 \pm 0.006$ & $-0.031 \pm 0.004$ & $-0.041 \pm 0.004$ \\
    \cmidrule(lr){2-5}
     & REVE-Base & $-0.014 \pm 0.006$ & $-0.012 \pm 0.004$ & $-0.016 \pm 0.004$ \\
     & REVE-Large & $\mathbf{-0.003} \pm 0.005$ & $\mathbf{-0.004} \pm 0.003$ & $\mathbf{-0.007} \pm 0.002$ \\
    \cmidrule(lr){1-5}
    \multirow{4}{*}{\vspace{-10pt}Tent} & CBraMod & $-0.248 \pm 0.005$ & $-0.271 \pm 0.042$ & $-0.324 \pm 0.028$ \\
    \cmidrule(lr){2-5}
     & TFM & $-0.196 \pm 0.036$ & $-0.183 \pm 0.034$ & $-0.174 \pm 0.042$ \\
    \cmidrule(lr){2-5}
     & REVE-Base & $-0.223 \pm 0.045$ & $-0.223 \pm 0.059$ & $-0.243 \pm 0.074$ \\
     & REVE-Large & $-0.166 \pm 0.084$ & $-0.134 \pm 0.076$ & $-0.147 \pm 0.088$ \\
    \bottomrule
  \end{tabular}
\end{table}

\begin{table}[H]
  \centering
  \scriptsize
  \setlength{\tabcolsep}{3pt}
  \renewcommand{\arraystretch}{1.05}
  \caption{CHB-MIT delta relative to the no-TTA baseline.}
  \label{tab:chbmit_delta}
  \begin{tabular}{p{1.45cm}p{1.75cm}ccc}
    \toprule
    \textbf{\shortstack{TTA\\Method}} & \textbf{\shortstack{Foundation\\Model}} & \textbf{Bal. Acc. $\Delta$} & \textbf{ROC AUC $\Delta$} & \textbf{PR AUC $\Delta$} \\
    \midrule
    \multirow{4}{*}{\vspace{-10pt}SHOT} & CBraMod & $+0.014 \pm 0.030$ & $-0.246 \pm 0.039$ & $-0.069 \pm 0.020$ \\
    \cmidrule(lr){2-5}
     & TFM & $+0.007 \pm 0.007$ & $-0.003 \pm 0.006$ & $-0.027 \pm 0.026$ \\
    \cmidrule(lr){2-5}
     & REVE-Base & $+0.004 \pm 0.059$ & $-0.325 \pm 0.134$ & $-0.287 \pm 0.026$ \\
     & REVE-Large & $-0.018 \pm 0.096$ & $-0.210 \pm 0.116$ & $-0.331 \pm 0.093$ \\
    \cmidrule(lr){1-5}
    \multirow{4}{*}{\vspace{-10pt}T3A} & CBraMod & $+0.000 \pm 0.000$ & $+0.002 \pm 0.006$ & $+0.001 \pm 0.005$ \\
    \cmidrule(lr){2-5}
     & TFM & $+0.067 \pm 0.006$ & $-0.141 \pm 0.019$ & $-0.078 \pm 0.017$ \\
    \cmidrule(lr){2-5}
     & REVE-Base & $\mathbf{+0.189} \pm 0.048$ & $\mathbf{+0.011} \pm 0.005$ & $-0.007 \pm 0.027$ \\
     & REVE-Large & $+0.187 \pm 0.035$ & $-0.001 \pm 0.008$ & $-0.023 \pm 0.065$ \\
    \cmidrule(lr){1-5}
    \multirow{4}{*}{\vspace{-10pt}Tent} & CBraMod & $+0.000 \pm 0.001$ & $-0.257 \pm 0.056$ & $-0.070 \pm 0.029$ \\
    \cmidrule(lr){2-5}
     & TFM & $-0.004 \pm 0.004$ & $-0.001 \pm 0.002$ & $\mathbf{+0.020} \pm 0.013$ \\
    \cmidrule(lr){2-5}
     & REVE-Base & $-0.013 \pm 0.016$ & $-0.002 \pm 0.002$ & $+0.009 \pm 0.007$ \\
     & REVE-Large & $-0.063 \pm 0.028$ & $-0.034 \pm 0.008$ & $-0.005 \pm 0.022$ \\
    \bottomrule
  \end{tabular}
\end{table}

\begin{table}[H]
  \centering
  \scriptsize
  \setlength{\tabcolsep}{3pt}
  \renewcommand{\arraystretch}{1.05}
  \caption{Sleep-EDF delta relative to the no-TTA baseline.}
  \label{tab:sleep_edf_78_delta}
  \begin{tabular}{p{1.45cm}p{1.75cm}ccc}
    \toprule
    \textbf{\shortstack{TTA\\Method}} & \textbf{\shortstack{Foundation\\Model}} & \textbf{Bal. Acc. $\Delta$} & \textbf{Cohen's $\kappa$ $\Delta$} & \textbf{Weighted F1 $\Delta$} \\
    \midrule
    \multirow{4}{*}{\vspace{-10pt}SHOT} & CBraMod & $-0.315 \pm 0.013$ & $-0.554 \pm 0.003$ & $-0.627 \pm 0.044$ \\
    \cmidrule(lr){2-5}
     & TFM & $+0.004 \pm 0.003$ & $\mathbf{-0.006} \pm 0.005$ & $\mathbf{-0.002} \pm 0.002$ \\
    \cmidrule(lr){2-5}
     & REVE-Base & $-0.315 \pm 0.060$ & $-0.511 \pm 0.075$ & $-0.461 \pm 0.101$ \\
     & REVE-Large & $-0.418 \pm 0.019$ & $-0.640 \pm 0.025$ & $-0.664 \pm 0.057$ \\
    \cmidrule(lr){1-5}
    \multirow{4}{*}{\vspace{-10pt}T3A} & CBraMod & $\mathbf{+0.022} \pm 0.009$ & $-0.154 \pm 0.013$ & $-0.093 \pm 0.012$ \\
    \cmidrule(lr){2-5}
     & TFM & $-0.014 \pm 0.007$ & $-0.130 \pm 0.020$ & $-0.076 \pm 0.017$ \\
    \cmidrule(lr){2-5}
     & REVE-Base & $-0.042 \pm 0.010$ & $-0.229 \pm 0.014$ & $-0.160 \pm 0.014$ \\
     & REVE-Large & $-0.026 \pm 0.006$ & $-0.215 \pm 0.015$ & $-0.144 \pm 0.014$ \\
    \cmidrule(lr){1-5}
    \multirow{4}{*}{\vspace{-10pt}Tent} & CBraMod & $-0.312 \pm 0.014$ & $-0.552 \pm 0.004$ & $-0.508 \pm 0.039$ \\
    \cmidrule(lr){2-5}
     & TFM & $-0.076 \pm 0.058$ & $-0.086 \pm 0.109$ & $-0.090 \pm 0.087$ \\
    \cmidrule(lr){2-5}
     & REVE-Base & $-0.183 \pm 0.090$ & $-0.176 \pm 0.173$ & $-0.160 \pm 0.172$ \\
     & REVE-Large & $-0.155 \pm 0.051$ & $-0.121 \pm 0.081$ & $-0.102 \pm 0.054$ \\
    \bottomrule
  \end{tabular}
\end{table}

\begin{table}[H]
  \centering
  \scriptsize
  \setlength{\tabcolsep}{3pt}
  \renewcommand{\arraystretch}{1.05}
  \caption{EAR-EEG delta relative to the no-TTA baseline.}
  \label{tab:eareeg_delta_2}
  \begin{tabular}{p{1.45cm}p{1.75cm}ccc}
    \toprule
    \textbf{\shortstack{TTA\\Method}} & \textbf{\shortstack{Foundation\\Model}} & \textbf{Bal. Acc. $\Delta$} & \textbf{Cohen's $\kappa$ $\Delta$} & \textbf{Weighted F1 $\Delta$} \\
    \midrule
    \multirow{4}{*}{\vspace{-10pt}SHOT} & CBraMod & $-0.065 \pm 0.020$ & $-0.087 \pm 0.027$ & $-0.224 \pm 0.015$ \\
    \cmidrule(lr){2-5}
     & TFM & $+0.000 \pm 0.000$ & $-0.000 \pm 0.001$ & $-0.000 \pm 0.000$ \\
    \cmidrule(lr){2-5}
     & REVE-Base & $-0.018 \pm 0.032$ & $-0.085 \pm 0.064$ & $-0.119 \pm 0.076$ \\
     & REVE-Large & $-0.068 \pm 0.056$ & $-0.156 \pm 0.083$ & $-0.150 \pm 0.087$ \\
    \cmidrule(lr){1-5}
    \multirow{4}{*}{\vspace{-10pt}T3A} & CBraMod & $\mathbf{+0.048} \pm 0.012$ & $\mathbf{+0.064} \pm 0.016$ & $\mathbf{+0.018} \pm 0.009$ \\
    \cmidrule(lr){2-5}
     & TFM & $-0.005 \pm 0.007$ & $-0.042 \pm 0.006$ & $-0.009 \pm 0.006$ \\
    \cmidrule(lr){2-5}
     & REVE-Base & $+0.037 \pm 0.007$ & $+0.001 \pm 0.015$ & $+0.001 \pm 0.017$ \\
     & REVE-Large & $+0.022 \pm 0.007$ & $-0.010 \pm 0.014$ & $-0.009 \pm 0.010$ \\
    \cmidrule(lr){1-5}
    \multirow{4}{*}{\vspace{-10pt}Tent} & CBraMod & $-0.064 \pm 0.022$ & $-0.084 \pm 0.030$ & $-0.189 \pm 0.060$ \\
    \cmidrule(lr){2-5}
     & TFM & $-0.001 \pm 0.001$ & $+0.000 \pm 0.001$ & $-0.000 \pm 0.001$ \\
    \cmidrule(lr){2-5}
     & REVE-Base & $-0.032 \pm 0.022$ & $-0.047 \pm 0.035$ & $-0.046 \pm 0.028$ \\
     & REVE-Large & $-0.018 \pm 0.014$ & $-0.025 \pm 0.018$ & $-0.019 \pm 0.015$ \\
    \bottomrule
  \end{tabular}
\end{table}

\subsection{Relative Performance Change}
\label{sec:appendix_percentage}

We report TTA-minus-No-TTA deltas together with the corresponding
relative change. Relative change is computed as $100\times\text{mean }\Delta /
\text{mean No-TTA}$. Values use the same seed and batch-size aggregation as
Appendix~\ref{sec:appendix_delta_tables}.

\begin{table}[H]
  \centering
  \scriptsize
  \setlength{\tabcolsep}{3pt}
  \renewcommand{\arraystretch}{1.05}
  \caption{TUEV relative performance change from No-TTA. Entries show mean delta $\pm$ standard deviation, with relative change in parentheses.}
  \label{tab:tuev_pct}
  \begin{tabular}{p{1.45cm}p{1.75cm}ccc}
    \toprule
    \textbf{\shortstack{TTA\\Method}} & \textbf{\shortstack{Foundation\\Model}} & \textbf{\shortstack{Bal. Acc.\\$\Delta$ (\%)}} & \textbf{\shortstack{Cohen's $\kappa$\\$\Delta$ (\%)}} & \textbf{\shortstack{Weighted F1\\$\Delta$ (\%)}} \\
    \midrule
    \multirow{4}{*}{\vspace{-10pt}SHOT} & CBraMod & $-0.210 \pm 0.025$~{\scriptsize$(-55\%)$} & $-0.462 \pm 0.056$~{\scriptsize$(-99\%)$} & $-0.673 \pm 0.016$~{\scriptsize$(-93\%)$} \\
    \cmidrule(lr){2-5}
     & TFM & $+0.024 \pm 0.014$~{\scriptsize$(+7\%)$} & $-0.001 \pm 0.022$~{\scriptsize$(-0\%)$} & $-0.010 \pm 0.018$~{\scriptsize$(-1\%)$} \\
    \cmidrule(lr){2-5}
     & REVE-Base & $-0.095 \pm 0.090$~{\scriptsize$(-21\%)$} & $-0.431 \pm 0.056$~{\scriptsize$(-77\%)$} & $-0.574 \pm 0.080$~{\scriptsize$(-74\%)$} \\
     & REVE-Large & $-0.137 \pm 0.035$~{\scriptsize$(-28\%)$} & $-0.473 \pm 0.037$~{\scriptsize$(-81\%)$} & $-0.624 \pm 0.050$~{\scriptsize$(-79\%)$} \\
    \cmidrule(lr){1-5}
    \multirow{4}{*}{\vspace{-10pt}T3A} & CBraMod & $+0.031 \pm 0.019$~{\scriptsize$(+8\%)$} & $-0.118 \pm 0.061$~{\scriptsize$(-26\%)$} & $-0.096 \pm 0.051$~{\scriptsize$(-13\%)$} \\
    \cmidrule(lr){2-5}
     & TFM & $+0.012 \pm 0.049$~{\scriptsize$(+3\%)$} & $-0.181 \pm 0.044$~{\scriptsize$(-45\%)$} & $-0.174 \pm 0.040$~{\scriptsize$(-25\%)$} \\
    \cmidrule(lr){2-5}
     & REVE-Base & $+0.072 \pm 0.042$~{\scriptsize$(+16\%)$} & $-0.137 \pm 0.076$~{\scriptsize$(-25\%)$} & $-0.095 \pm 0.074$~{\scriptsize$(-12\%)$} \\
     & REVE-Large & $+0.095 \pm 0.050$~{\scriptsize$(+19\%)$} & $-0.081 \pm 0.037$~{\scriptsize$(-14\%)$} & $-0.044 \pm 0.022$~{\scriptsize$(-6\%)$} \\
    \cmidrule(lr){1-5}
    \multirow{4}{*}{\vspace{-10pt}Tent} & CBraMod & $-0.212 \pm 0.026$~{\scriptsize$(-56\%)$} & $-0.465 \pm 0.056$~{\scriptsize$(-100\%)$} & $-0.253 \pm 0.179$~{\scriptsize$(-35\%)$} \\
    \cmidrule(lr){2-5}
     & TFM & $-0.101 \pm 0.043$~{\scriptsize$(-27\%)$} & $-0.169 \pm 0.095$~{\scriptsize$(-42\%)$} & $-0.068 \pm 0.038$~{\scriptsize$(-10\%)$} \\
    \cmidrule(lr){2-5}
     & REVE-Base & $-0.250 \pm 0.029$~{\scriptsize$(-55\%)$} & $-0.474 \pm 0.075$~{\scriptsize$(-85\%)$} & $-0.198 \pm 0.033$~{\scriptsize$(-26\%)$} \\
     & REVE-Large & $-0.208 \pm 0.061$~{\scriptsize$(-42\%)$} & $-0.328 \pm 0.119$~{\scriptsize$(-56\%)$} & $-0.140 \pm 0.050$~{\scriptsize$(-18\%)$} \\
    \bottomrule
  \end{tabular}
\end{table}

\begin{table}[H]
  \centering
  \scriptsize
  \setlength{\tabcolsep}{3pt}
  \renewcommand{\arraystretch}{1.05}
  \caption{TUAB relative performance change from No-TTA. Entries show mean delta $\pm$ standard deviation, with relative change in parentheses.}
  \label{tab:tuab_pct}
  \begin{tabular}{p{1.45cm}p{1.75cm}ccc}
    \toprule
    \textbf{\shortstack{TTA\\Method}} & \textbf{\shortstack{Foundation\\Model}} & \textbf{\shortstack{Bal. Acc.\\$\Delta$ (\%)}} & \textbf{\shortstack{ROC AUC\\$\Delta$ (\%)}} & \textbf{\shortstack{PR AUC\\$\Delta$ (\%)}} \\
    \midrule
    \multirow{4}{*}{\vspace{-10pt}SHOT} & CBraMod & $-0.247 \pm 0.005$~{\scriptsize$(-33\%)$} & $-0.303 \pm 0.049$~{\scriptsize$(-37\%)$} & $-0.337 \pm 0.032$~{\scriptsize$(-41\%)$} \\
    \cmidrule(lr){2-5}
     & TFM & $-0.030 \pm 0.016$~{\scriptsize$(-4\%)$} & $-0.012 \pm 0.008$~{\scriptsize$(-1\%)$} & $-0.011 \pm 0.007$~{\scriptsize$(-1\%)$} \\
    \cmidrule(lr){2-5}
     & REVE-Base & $-0.147 \pm 0.058$~{\scriptsize$(-18\%)$} & $-0.090 \pm 0.059$~{\scriptsize$(-10\%)$} & $-0.097 \pm 0.065$~{\scriptsize$(-11\%)$} \\
     & REVE-Large & $-0.110 \pm 0.055$~{\scriptsize$(-14\%)$} & $-0.056 \pm 0.030$~{\scriptsize$(-6\%)$} & $-0.060 \pm 0.032$~{\scriptsize$(-7\%)$} \\
    \cmidrule(lr){1-5}
    \multirow{4}{*}{\vspace{-10pt}T3A} & CBraMod & $-0.025 \pm 0.004$~{\scriptsize$(-3\%)$} & $-0.021 \pm 0.004$~{\scriptsize$(-3\%)$} & $-0.029 \pm 0.007$~{\scriptsize$(-4\%)$} \\
    \cmidrule(lr){2-5}
     & TFM & $-0.027 \pm 0.006$~{\scriptsize$(-4\%)$} & $-0.031 \pm 0.004$~{\scriptsize$(-4\%)$} & $-0.041 \pm 0.004$~{\scriptsize$(-5\%)$} \\
    \cmidrule(lr){2-5}
     & REVE-Base & $-0.014 \pm 0.006$~{\scriptsize$(-2\%)$} & $-0.012 \pm 0.004$~{\scriptsize$(-1\%)$} & $-0.016 \pm 0.004$~{\scriptsize$(-2\%)$} \\
     & REVE-Large & $-0.003 \pm 0.005$~{\scriptsize$(-0\%)$} & $-0.004 \pm 0.003$~{\scriptsize$(-0\%)$} & $-0.007 \pm 0.002$~{\scriptsize$(-1\%)$} \\
    \cmidrule(lr){1-5}
    \multirow{4}{*}{\vspace{-10pt}Tent} & CBraMod & $-0.248 \pm 0.005$~{\scriptsize$(-33\%)$} & $-0.271 \pm 0.042$~{\scriptsize$(-33\%)$} & $-0.324 \pm 0.028$~{\scriptsize$(-39\%)$} \\
    \cmidrule(lr){2-5}
     & TFM & $-0.196 \pm 0.036$~{\scriptsize$(-26\%)$} & $-0.183 \pm 0.034$~{\scriptsize$(-22\%)$} & $-0.174 \pm 0.042$~{\scriptsize$(-21\%)$} \\
    \cmidrule(lr){2-5}
     & REVE-Base & $-0.223 \pm 0.045$~{\scriptsize$(-28\%)$} & $-0.223 \pm 0.059$~{\scriptsize$(-25\%)$} & $-0.243 \pm 0.074$~{\scriptsize$(-27\%)$} \\
     & REVE-Large & $-0.166 \pm 0.084$~{\scriptsize$(-21\%)$} & $-0.134 \pm 0.076$~{\scriptsize$(-15\%)$} & $-0.147 \pm 0.088$~{\scriptsize$(-16\%)$} \\
    \bottomrule
  \end{tabular}
\end{table}

\begin{table}[H]
  \centering
  \scriptsize
  \setlength{\tabcolsep}{3pt}
  \renewcommand{\arraystretch}{1.05}
  \caption{CHB-MIT relative performance change from No-TTA. Entries show mean delta $\pm$ standard deviation, with relative change in parentheses.}
  \label{tab:chbmit_pct}
  \begin{tabular}{p{1.45cm}p{1.75cm}ccc}
    \toprule
    \textbf{\shortstack{TTA\\Method}} & \textbf{\shortstack{Foundation\\Model}} & \textbf{\shortstack{Bal. Acc.\\$\Delta$ (\%)}} & \textbf{\shortstack{ROC AUC\\$\Delta$ (\%)}} & \textbf{\shortstack{PR AUC\\$\Delta$ (\%)}} \\
    \midrule
    \multirow{4}{*}{\vspace{-10pt}SHOT} & CBraMod & $+0.014 \pm 0.030$~{\scriptsize$(+3\%)$} & $-0.246 \pm 0.039$~{\scriptsize$(-33\%)$} & $-0.069 \pm 0.020$~{\scriptsize$(-75\%)$} \\
    \cmidrule(lr){2-5}
     & TFM & $+0.007 \pm 0.007$~{\scriptsize$(+1\%)$} & $-0.003 \pm 0.006$~{\scriptsize$(-0\%)$} & $-0.027 \pm 0.026$~{\scriptsize$(-8\%)$} \\
    \cmidrule(lr){2-5}
     & REVE-Base & $+0.004 \pm 0.059$~{\scriptsize$(+1\%)$} & $-0.325 \pm 0.134$~{\scriptsize$(-39\%)$} & $-0.287 \pm 0.026$~{\scriptsize$(-90\%)$} \\
     & REVE-Large & $-0.018 \pm 0.096$~{\scriptsize$(-3\%)$} & $-0.210 \pm 0.116$~{\scriptsize$(-24\%)$} & $-0.331 \pm 0.093$~{\scriptsize$(-82\%)$} \\
    \cmidrule(lr){1-5}
    \multirow{4}{*}{\vspace{-10pt}T3A} & CBraMod & $+0.000 \pm 0.000$~{\scriptsize$(+0\%)$} & $+0.002 \pm 0.006$~{\scriptsize$(+0\%)$} & $+0.001 \pm 0.005$~{\scriptsize$(+1\%)$} \\
    \cmidrule(lr){2-5}
     & TFM & $+0.067 \pm 0.006$~{\scriptsize$(+12\%)$} & $-0.141 \pm 0.019$~{\scriptsize$(-16\%)$} & $-0.078 \pm 0.017$~{\scriptsize$(-24\%)$} \\
    \cmidrule(lr){2-5}
     & REVE-Base & $+0.189 \pm 0.048$~{\scriptsize$(+34\%)$} & $+0.011 \pm 0.005$~{\scriptsize$(+1\%)$} & $-0.007 \pm 0.027$~{\scriptsize$(-2\%)$} \\
     & REVE-Large & $+0.187 \pm 0.035$~{\scriptsize$(+31\%)$} & $-0.001 \pm 0.008$~{\scriptsize$(-0\%)$} & $-0.023 \pm 0.065$~{\scriptsize$(-6\%)$} \\
    \cmidrule(lr){1-5}
    \multirow{4}{*}{\vspace{-10pt}Tent} & CBraMod & $+0.000 \pm 0.001$~{\scriptsize$(+0\%)$} & $-0.257 \pm 0.056$~{\scriptsize$(-34\%)$} & $-0.070 \pm 0.029$~{\scriptsize$(-75\%)$} \\
    \cmidrule(lr){2-5}
     & TFM & $-0.004 \pm 0.004$~{\scriptsize$(-1\%)$} & $-0.001 \pm 0.002$~{\scriptsize$(-0\%)$} & $+0.020 \pm 0.013$~{\scriptsize$(+6\%)$} \\
    \cmidrule(lr){2-5}
     & REVE-Base & $-0.013 \pm 0.016$~{\scriptsize$(-2\%)$} & $-0.002 \pm 0.002$~{\scriptsize$(-0\%)$} & $+0.009 \pm 0.007$~{\scriptsize$(+3\%)$} \\
     & REVE-Large & $-0.063 \pm 0.028$~{\scriptsize$(-10\%)$} & $-0.034 \pm 0.008$~{\scriptsize$(-4\%)$} & $-0.005 \pm 0.022$~{\scriptsize$(-1\%)$} \\
    \bottomrule
  \end{tabular}
\end{table}

\begin{table}[H]
  \centering
  \scriptsize
  \setlength{\tabcolsep}{3pt}
  \renewcommand{\arraystretch}{1.05}
  \caption{Sleep-EDF relative performance change from No-TTA. Entries show mean delta $\pm$ standard deviation, with relative change in parentheses.}
  \label{tab:sleep_edf_78_pct}
  \begin{tabular}{p{1.45cm}p{1.75cm}ccc}
    \toprule
    \textbf{\shortstack{TTA\\Method}} & \textbf{\shortstack{Foundation\\Model}} & \textbf{\shortstack{Bal. Acc.\\$\Delta$ (\%)}} & \textbf{\shortstack{Cohen's $\kappa$\\$\Delta$ (\%)}} & \textbf{\shortstack{Weighted F1\\$\Delta$ (\%)}} \\
    \midrule
    \multirow{4}{*}{\vspace{-10pt}SHOT} & CBraMod & $-0.315 \pm 0.013$~{\scriptsize$(-61\%)$} & $-0.554 \pm 0.003$~{\scriptsize$(-100\%)$} & $-0.627 \pm 0.044$~{\scriptsize$(-95\%)$} \\
    \cmidrule(lr){2-5}
     & TFM & $+0.004 \pm 0.003$~{\scriptsize$(+1\%)$} & $-0.006 \pm 0.005$~{\scriptsize$(-1\%)$} & $-0.002 \pm 0.002$~{\scriptsize$(-0\%)$} \\
    \cmidrule(lr){2-5}
     & REVE-Base & $-0.315 \pm 0.060$~{\scriptsize$(-51\%)$} & $-0.511 \pm 0.075$~{\scriptsize$(-80\%)$} & $-0.461 \pm 0.101$~{\scriptsize$(-63\%)$} \\
     & REVE-Large & $-0.418 \pm 0.019$~{\scriptsize$(-64\%)$} & $-0.640 \pm 0.025$~{\scriptsize$(-94\%)$} & $-0.664 \pm 0.057$~{\scriptsize$(-87\%)$} \\
    \cmidrule(lr){1-5}
    \multirow{4}{*}{\vspace{-10pt}T3A} & CBraMod & $+0.022 \pm 0.009$~{\scriptsize$(+4\%)$} & $-0.154 \pm 0.013$~{\scriptsize$(-28\%)$} & $-0.093 \pm 0.012$~{\scriptsize$(-14\%)$} \\
    \cmidrule(lr){2-5}
     & TFM & $-0.014 \pm 0.007$~{\scriptsize$(-2\%)$} & $-0.130 \pm 0.020$~{\scriptsize$(-23\%)$} & $-0.076 \pm 0.017$~{\scriptsize$(-11\%)$} \\
    \cmidrule(lr){2-5}
     & REVE-Base & $-0.042 \pm 0.010$~{\scriptsize$(-7\%)$} & $-0.229 \pm 0.014$~{\scriptsize$(-36\%)$} & $-0.160 \pm 0.014$~{\scriptsize$(-22\%)$} \\
     & REVE-Large & $-0.026 \pm 0.006$~{\scriptsize$(-4\%)$} & $-0.215 \pm 0.015$~{\scriptsize$(-32\%)$} & $-0.144 \pm 0.014$~{\scriptsize$(-19\%)$} \\
    \cmidrule(lr){1-5}
    \multirow{4}{*}{\vspace{-10pt}Tent} & CBraMod & $-0.312 \pm 0.014$~{\scriptsize$(-61\%)$} & $-0.552 \pm 0.004$~{\scriptsize$(-100\%)$} & $-0.508 \pm 0.039$~{\scriptsize$(-77\%)$} \\
    \cmidrule(lr){2-5}
     & TFM & $-0.076 \pm 0.058$~{\scriptsize$(-14\%)$} & $-0.086 \pm 0.109$~{\scriptsize$(-15\%)$} & $-0.090 \pm 0.087$~{\scriptsize$(-13\%)$} \\
    \cmidrule(lr){2-5}
     & REVE-Base & $-0.183 \pm 0.090$~{\scriptsize$(-29\%)$} & $-0.176 \pm 0.173$~{\scriptsize$(-28\%)$} & $-0.160 \pm 0.172$~{\scriptsize$(-22\%)$} \\
     & REVE-Large & $-0.155 \pm 0.051$~{\scriptsize$(-24\%)$} & $-0.121 \pm 0.081$~{\scriptsize$(-18\%)$} & $-0.102 \pm 0.054$~{\scriptsize$(-13\%)$} \\
    \bottomrule
  \end{tabular}
\end{table}

\begin{table}[H]
  \centering
  \scriptsize
  \setlength{\tabcolsep}{3pt}
  \renewcommand{\arraystretch}{1.05}
  \caption{EAR-EEG relative performance change from No-TTA. Entries show mean delta $\pm$ standard deviation, with relative change in parentheses.}
  \label{tab:eareeg_pct}
  \begin{tabular}{p{1.45cm}p{1.75cm}ccc}
    \toprule
    \textbf{\shortstack{TTA\\Method}} & \textbf{\shortstack{Foundation\\Model}} & \textbf{\shortstack{Bal. Acc.\\$\Delta$ (\%)}} & \textbf{\shortstack{Cohen's $\kappa$\\$\Delta$ (\%)}} & \textbf{\shortstack{Weighted F1\\$\Delta$ (\%)}} \\
    \midrule
    \multirow{4}{*}{\vspace{-10pt}SHOT} & CBraMod & $-0.065 \pm 0.020$~{\scriptsize$(-27\%)$} & $-0.087 \pm 0.027$~{\scriptsize$(-93\%)$} & $-0.224 \pm 0.015$~{\scriptsize$(-75\%)$} \\
    \cmidrule(lr){2-5}
     & TFM & $+0.000 \pm 0.000$~{\scriptsize$(+0\%)$} & $-0.000 \pm 0.001$~{\scriptsize$(-0\%)$} & $-0.000 \pm 0.000$~{\scriptsize$(-0\%)$} \\
    \cmidrule(lr){2-5}
     & REVE-Base & $-0.018 \pm 0.032$~{\scriptsize$(-5\%)$} & $-0.085 \pm 0.064$~{\scriptsize$(-28\%)$} & $-0.119 \pm 0.076$~{\scriptsize$(-25\%)$} \\
     & REVE-Large & $-0.068 \pm 0.056$~{\scriptsize$(-17\%)$} & $-0.156 \pm 0.083$~{\scriptsize$(-42\%)$} & $-0.150 \pm 0.087$~{\scriptsize$(-29\%)$} \\
    \cmidrule(lr){1-5}
    \multirow{4}{*}{\vspace{-10pt}T3A} & CBraMod & $+0.048 \pm 0.012$~{\scriptsize$(+20\%)$} & $+0.064 \pm 0.016$~{\scriptsize$(+69\%)$} & $+0.018 \pm 0.009$~{\scriptsize$(+6\%)$} \\
    \cmidrule(lr){2-5}
     & TFM & $-0.005 \pm 0.007$~{\scriptsize$(-1\%)$} & $-0.042 \pm 0.006$~{\scriptsize$(-15\%)$} & $-0.009 \pm 0.006$~{\scriptsize$(-2\%)$} \\
    \cmidrule(lr){2-5}
     & REVE-Base & $+0.037 \pm 0.007$~{\scriptsize$(+10\%)$} & $+0.001 \pm 0.015$~{\scriptsize$(+0\%)$} & $+0.001 \pm 0.017$~{\scriptsize$(+0\%)$} \\
     & REVE-Large & $+0.022 \pm 0.007$~{\scriptsize$(+6\%)$} & $-0.010 \pm 0.014$~{\scriptsize$(-3\%)$} & $-0.009 \pm 0.010$~{\scriptsize$(-2\%)$} \\
    \cmidrule(lr){1-5}
    \multirow{4}{*}{\vspace{-10pt}Tent} & CBraMod & $-0.064 \pm 0.022$~{\scriptsize$(-27\%)$} & $-0.084 \pm 0.030$~{\scriptsize$(-90\%)$} & $-0.189 \pm 0.060$~{\scriptsize$(-63\%)$} \\
    \cmidrule(lr){2-5}
     & TFM & $-0.001 \pm 0.001$~{\scriptsize$(-0\%)$} & $+0.000 \pm 0.001$~{\scriptsize$(+0\%)$} & $-0.000 \pm 0.001$~{\scriptsize$(-0\%)$} \\
    \cmidrule(lr){2-5}
     & REVE-Base & $-0.032 \pm 0.022$~{\scriptsize$(-9\%)$} & $-0.047 \pm 0.035$~{\scriptsize$(-15\%)$} & $-0.046 \pm 0.028$~{\scriptsize$(-10\%)$} \\
     & REVE-Large & $-0.018 \pm 0.014$~{\scriptsize$(-5\%)$} & $-0.025 \pm 0.018$~{\scriptsize$(-7\%)$} & $-0.019 \pm 0.015$~{\scriptsize$(-4\%)$} \\
    \bottomrule
  \end{tabular}
\end{table}

\subsection{Statistical Significance Testing}
\label{sec_appendix_significance}

We compare each TTA method with the matched No-TTA baseline for every
method, foundation model, dataset, and metric. Deltas are computed within
each (seed, batch size) run as $\Delta_{\text{TTA}} =
\text{metric}_{\text{TTA}} - \text{metric}_{\text{No-TTA}}$. Because batch sizes
reuse the same seed/split/model, we average the three batch-size deltas
within each seed, yielding $5$ seed-level deltas for each of $180$
comparisons. We run two-sided paired $t$-tests against zero and apply
Benjamini--Hochberg FDR correction over all $180$ tests.
Tables~\ref{tab_tuev_significance}--\ref{tab_eareeg_significance}
report the FDR-adjusted $q$-values by dataset.
After correction, 145 comparisons are significant ($q<0.05$), with
17 improvements and 128 degradations. Significant effects are
mostly harmful, SHOT and Tent mainly degrade performance, and T3A accounts
for most improvements while remaining mixed across settings.

\begin{table}[H]
  \centering
  \scriptsize
  \setlength{\tabcolsep}{3pt}
  \renewcommand{\arraystretch}{1.05}
  \caption{TUEV paired seed-level $t$-tests comparing each TTA method against No-TTA, with Benjamini--Hochberg FDR correction. Asterisks denote FDR-adjusted significance levels, with \textsuperscript{*}$q<0.05$, \textsuperscript{**}$q<0.01$, and \textsuperscript{***}$q<0.001$. Unmarked comparisons are not significant.}
  \label{tab_tuev_significance}
  \begin{adjustbox}{max width=\textwidth}
  \begin{tabular}{p{1.35cm}p{1.55cm}p{1.75cm}ccc}
    \toprule
    \textbf{\shortstack{TTA\\Method}} & \textbf{\shortstack{Foundation\\Model}} & \textbf{Metric} & \textbf{Mean $\Delta$} & \textbf{$95\%$ CI} & \textbf{$q$} \\
    \midrule
    \multirow{12}{*}{\vspace{-15pt}SHOT} & \multirow{3}{*}{CBraMod} & Bal. Acc. & $-0.210$ & $[-0.243, -0.176]$ & $<0.001$\textsuperscript{***} \\
     &  & Cohen's $\kappa$ & $-0.462$ & $[-0.537, -0.387]$ & $<0.001$\textsuperscript{***} \\
     &  & Weighted F1 & $-0.673$ & $[-0.693, -0.653]$ & $<0.001$\textsuperscript{***} \\
    \cmidrule(lr){2-6}
     & \multirow{3}{*}{TFM} & Bal. Acc. & $+0.024$ & $[+0.008, +0.041]$ & $0.020$\textsuperscript{*} \\
     &  & Cohen's $\kappa$ & $-0.001$ & $[-0.021, +0.018]$ & $0.881$ \\
     &  & Weighted F1 & $-0.010$ & $[-0.025, +0.004]$ & $0.142$ \\
    \cmidrule(lr){2-6}
     & \multirow{3}{*}{REVE-Base} & Bal. Acc. & $-0.095$ & $[-0.190, +0.001]$ & $0.063$ \\
     &  & Cohen's $\kappa$ & $-0.431$ & $[-0.482, -0.380]$ & $<0.001$\textsuperscript{***} \\
     &  & Weighted F1 & $-0.574$ & $[-0.630, -0.517]$ & $<0.001$\textsuperscript{***} \\
     & \multirow{3}{*}{REVE-Large} & Bal. Acc. & $-0.137$ & $[-0.170, -0.104]$ & $<0.001$\textsuperscript{***} \\
     &  & Cohen's $\kappa$ & $-0.473$ & $[-0.516, -0.430]$ & $<0.001$\textsuperscript{***} \\
     &  & Weighted F1 & $-0.624$ & $[-0.676, -0.573]$ & $<0.001$\textsuperscript{***} \\
    \cmidrule(lr){1-6}
    \multirow{12}{*}{\vspace{-15pt}T3A} & \multirow{3}{*}{CBraMod} & Bal. Acc. & $+0.031$ & $[+0.005, +0.056]$ & $0.036$\textsuperscript{*} \\
     &  & Cohen's $\kappa$ & $-0.118$ & $[-0.200, -0.037]$ & $0.022$\textsuperscript{*} \\
     &  & Weighted F1 & $-0.096$ & $[-0.165, -0.028]$ & $0.024$\textsuperscript{*} \\
    \cmidrule(lr){2-6}
     & \multirow{3}{*}{TFM} & Bal. Acc. & $+0.012$ & $[-0.054, +0.077]$ & $0.677$ \\
     &  & Cohen's $\kappa$ & $-0.181$ & $[-0.239, -0.122]$ & $0.002$\textsuperscript{**} \\
     &  & Weighted F1 & $-0.174$ & $[-0.228, -0.120]$ & $0.002$\textsuperscript{**} \\
    \cmidrule(lr){2-6}
     & \multirow{3}{*}{REVE-Base} & Bal. Acc. & $+0.072$ & $[+0.016, +0.128]$ & $0.030$\textsuperscript{*} \\
     &  & Cohen's $\kappa$ & $-0.137$ & $[-0.239, -0.035]$ & $0.027$\textsuperscript{*} \\
     &  & Weighted F1 & $-0.095$ & $[-0.194, +0.004]$ & $0.068$ \\
     & \multirow{3}{*}{REVE-Large} & Bal. Acc. & $+0.095$ & $[+0.028, +0.162]$ & $0.024$\textsuperscript{*} \\
     &  & Cohen's $\kappa$ & $-0.081$ & $[-0.130, -0.032]$ & $0.015$\textsuperscript{*} \\
     &  & Weighted F1 & $-0.044$ & $[-0.074, -0.014]$ & $0.021$\textsuperscript{*} \\
    \cmidrule(lr){1-6}
    \multirow{12}{*}{\vspace{-15pt}Tent} & \multirow{3}{*}{CBraMod} & Bal. Acc. & $-0.212$ & $[-0.246, -0.178]$ & $<0.001$\textsuperscript{***} \\
     &  & Cohen's $\kappa$ & $-0.465$ & $[-0.539, -0.390]$ & $<0.001$\textsuperscript{***} \\
     &  & Weighted F1 & $-0.253$ & $[-0.442, -0.063]$ & $0.027$\textsuperscript{*} \\
    \cmidrule(lr){2-6}
     & \multirow{3}{*}{TFM} & Bal. Acc. & $-0.101$ & $[-0.135, -0.066]$ & $0.002$\textsuperscript{**} \\
     &  & Cohen's $\kappa$ & $-0.169$ & $[-0.250, -0.088]$ & $0.008$\textsuperscript{**} \\
     &  & Weighted F1 & $-0.068$ & $[-0.106, -0.031]$ & $0.012$\textsuperscript{*} \\
    \cmidrule(lr){2-6}
     & \multirow{3}{*}{REVE-Base} & Bal. Acc. & $-0.250$ & $[-0.270, -0.231]$ & $<0.001$\textsuperscript{***} \\
     &  & Cohen's $\kappa$ & $-0.474$ & $[-0.522, -0.427]$ & $<0.001$\textsuperscript{***} \\
     &  & Weighted F1 & $-0.198$ & $[-0.221, -0.176]$ & $<0.001$\textsuperscript{***} \\
     & \multirow{3}{*}{REVE-Large} & Bal. Acc. & $-0.208$ & $[-0.229, -0.187]$ & $<0.001$\textsuperscript{***} \\
     &  & Cohen's $\kappa$ & $-0.328$ & $[-0.350, -0.305]$ & $<0.001$\textsuperscript{***} \\
     &  & Weighted F1 & $-0.140$ & $[-0.150, -0.130]$ & $<0.001$\textsuperscript{***} \\
    \bottomrule
  \end{tabular}
  \end{adjustbox}
\end{table}

\begin{table}[H]
  \centering
  \scriptsize
  \setlength{\tabcolsep}{3pt}
  \renewcommand{\arraystretch}{1.05}
  \caption{TUAB paired seed-level $t$-tests comparing each TTA method against No-TTA, with Benjamini--Hochberg FDR correction. Asterisks denote FDR-adjusted significance levels, with \textsuperscript{*}$q<0.05$, \textsuperscript{**}$q<0.01$, and \textsuperscript{***}$q<0.001$. Unmarked comparisons are not significant.}
  \label{tab_tuab_significance}
  \begin{adjustbox}{max width=\textwidth}
  \begin{tabular}{p{1.35cm}p{1.55cm}p{1.75cm}ccc}
    \toprule
    \textbf{\shortstack{TTA\\Method}} & \textbf{\shortstack{Foundation\\Model}} & \textbf{Metric} & \textbf{Mean $\Delta$} & \textbf{$95\%$ CI} & \textbf{$q$} \\
    \midrule
    \multirow{12}{*}{\vspace{-15pt}SHOT} & \multirow{3}{*}{CBraMod} & Bal. Acc. & $-0.247$ & $[-0.251, -0.242]$ & $<0.001$\textsuperscript{***} \\
     &  & ROC AUC & $-0.303$ & $[-0.362, -0.244]$ & $<0.001$\textsuperscript{***} \\
     &  & PR AUC & $-0.337$ & $[-0.372, -0.302]$ & $<0.001$\textsuperscript{***} \\
    \cmidrule(lr){2-6}
     & \multirow{3}{*}{TFM} & Bal. Acc. & $-0.030$ & $[-0.048, -0.012]$ & $0.015$\textsuperscript{*} \\
     &  & ROC AUC & $-0.012$ & $[-0.020, -0.004]$ & $0.018$\textsuperscript{*} \\
     &  & PR AUC & $-0.011$ & $[-0.018, -0.005]$ & $0.012$\textsuperscript{*} \\
    \cmidrule(lr){2-6}
     & \multirow{3}{*}{REVE-Base} & Bal. Acc. & $-0.147$ & $[-0.189, -0.105]$ & $0.002$\textsuperscript{**} \\
     &  & ROC AUC & $-0.090$ & $[-0.135, -0.044]$ & $0.009$\textsuperscript{**} \\
     &  & PR AUC & $-0.097$ & $[-0.148, -0.047]$ & $0.010$\textsuperscript{**} \\
     & \multirow{3}{*}{REVE-Large} & Bal. Acc. & $-0.110$ & $[-0.171, -0.049]$ & $0.012$\textsuperscript{*} \\
     &  & ROC AUC & $-0.056$ & $[-0.080, -0.032]$ & $0.005$\textsuperscript{**} \\
     &  & PR AUC & $-0.060$ & $[-0.084, -0.037]$ & $0.004$\textsuperscript{**} \\
    \cmidrule(lr){1-6}
    \multirow{12}{*}{\vspace{-15pt}T3A} & \multirow{3}{*}{CBraMod} & Bal. Acc. & $-0.025$ & $[-0.031, -0.019]$ & $<0.001$\textsuperscript{***} \\
     &  & ROC AUC & $-0.021$ & $[-0.026, -0.015]$ & $<0.001$\textsuperscript{***} \\
     &  & PR AUC & $-0.029$ & $[-0.039, -0.020]$ & $0.002$\textsuperscript{**} \\
    \cmidrule(lr){2-6}
     & \multirow{3}{*}{TFM} & Bal. Acc. & $-0.027$ & $[-0.035, -0.018]$ & $0.002$\textsuperscript{**} \\
     &  & ROC AUC & $-0.031$ & $[-0.036, -0.026]$ & $<0.001$\textsuperscript{***} \\
     &  & PR AUC & $-0.041$ & $[-0.047, -0.036]$ & $<0.001$\textsuperscript{***} \\
    \cmidrule(lr){2-6}
     & \multirow{3}{*}{REVE-Base} & Bal. Acc. & $-0.014$ & $[-0.023, -0.005]$ & $0.016$\textsuperscript{*} \\
     &  & ROC AUC & $-0.012$ & $[-0.017, -0.007]$ & $0.004$\textsuperscript{**} \\
     &  & PR AUC & $-0.016$ & $[-0.021, -0.010]$ & $0.003$\textsuperscript{**} \\
     & \multirow{3}{*}{REVE-Large} & Bal. Acc. & $-0.003$ & $[-0.010, +0.003]$ & $0.268$ \\
     &  & ROC AUC & $-0.004$ & $[-0.008, +0.001]$ & $0.110$ \\
     &  & PR AUC & $-0.007$ & $[-0.010, -0.005]$ & $0.003$\textsuperscript{**} \\
    \cmidrule(lr){1-6}
    \multirow{12}{*}{\vspace{-15pt}Tent} & \multirow{3}{*}{CBraMod} & Bal. Acc. & $-0.248$ & $[-0.254, -0.241]$ & $<0.001$\textsuperscript{***} \\
     &  & ROC AUC & $-0.271$ & $[-0.303, -0.239]$ & $<0.001$\textsuperscript{***} \\
     &  & PR AUC & $-0.324$ & $[-0.339, -0.308]$ & $<0.001$\textsuperscript{***} \\
    \cmidrule(lr){2-6}
     & \multirow{3}{*}{TFM} & Bal. Acc. & $-0.196$ & $[-0.220, -0.172]$ & $<0.001$\textsuperscript{***} \\
     &  & ROC AUC & $-0.183$ & $[-0.210, -0.156]$ & $<0.001$\textsuperscript{***} \\
     &  & PR AUC & $-0.174$ & $[-0.205, -0.142]$ & $<0.001$\textsuperscript{***} \\
    \cmidrule(lr){2-6}
     & \multirow{3}{*}{REVE-Base} & Bal. Acc. & $-0.223$ & $[-0.244, -0.202]$ & $<0.001$\textsuperscript{***} \\
     &  & ROC AUC & $-0.223$ & $[-0.256, -0.190]$ & $<0.001$\textsuperscript{***} \\
     &  & PR AUC & $-0.243$ & $[-0.273, -0.214]$ & $<0.001$\textsuperscript{***} \\
     & \multirow{3}{*}{REVE-Large} & Bal. Acc. & $-0.166$ & $[-0.205, -0.128]$ & $<0.001$\textsuperscript{***} \\
     &  & ROC AUC & $-0.134$ & $[-0.165, -0.103]$ & $<0.001$\textsuperscript{***} \\
     &  & PR AUC & $-0.147$ & $[-0.182, -0.113]$ & $<0.001$\textsuperscript{***} \\
    \bottomrule
  \end{tabular}
  \end{adjustbox}
\end{table}

\begin{table}[H]
  \centering
  \scriptsize
  \setlength{\tabcolsep}{3pt}
  \renewcommand{\arraystretch}{1.05}
  \caption{CHB-MIT paired seed-level $t$-tests comparing each TTA method against No-TTA, with Benjamini--Hochberg FDR correction. Asterisks denote FDR-adjusted significance levels, with \textsuperscript{*}$q<0.05$, \textsuperscript{**}$q<0.01$, and \textsuperscript{***}$q<0.001$. Unmarked comparisons are not significant.}
  \label{tab_chbmit_significance}
  \begin{adjustbox}{max width=\textwidth}
  \begin{tabular}{p{1.35cm}p{1.55cm}p{1.75cm}ccc}
    \toprule
    \textbf{\shortstack{TTA\\Method}} & \textbf{\shortstack{Foundation\\Model}} & \textbf{Metric} & \textbf{Mean $\Delta$} & \textbf{$95\%$ CI} & \textbf{$q$} \\
    \midrule
    \multirow{12}{*}{\vspace{-15pt}SHOT} & \multirow{3}{*}{CBraMod} & Bal. Acc. & $+0.014$ & $[-0.009, +0.036]$ & $0.192$ \\
     &  & ROC AUC & $-0.246$ & $[-0.291, -0.201]$ & $<0.001$\textsuperscript{***} \\
     &  & PR AUC & $-0.069$ & $[-0.092, -0.046]$ & $0.002$\textsuperscript{**} \\
    \cmidrule(lr){2-6}
     & \multirow{3}{*}{TFM} & Bal. Acc. & $+0.007$ & $[-0.002, +0.015]$ & $0.110$ \\
     &  & ROC AUC & $-0.003$ & $[-0.009, +0.003]$ & $0.248$ \\
     &  & PR AUC & $-0.027$ & $[-0.052, -0.003]$ & $0.044$\textsuperscript{*} \\
    \cmidrule(lr){2-6}
     & \multirow{3}{*}{REVE-Base} & Bal. Acc. & $+0.004$ & $[-0.073, +0.080]$ & $0.909$ \\
     &  & ROC AUC & $-0.325$ & $[-0.497, -0.152]$ & $0.010$\textsuperscript{*} \\
     &  & PR AUC & $-0.287$ & $[-0.320, -0.254]$ & $<0.001$\textsuperscript{***} \\
     & \multirow{3}{*}{REVE-Large} & Bal. Acc. & $-0.018$ & $[-0.142, +0.106]$ & $0.739$ \\
     &  & ROC AUC & $-0.210$ & $[-0.355, -0.064]$ & $0.022$\textsuperscript{*} \\
     &  & PR AUC & $-0.331$ & $[-0.454, -0.208]$ & $0.003$\textsuperscript{**} \\
    \cmidrule(lr){1-6}
    \multirow{12}{*}{\vspace{-15pt}T3A} & \multirow{3}{*}{CBraMod} & Bal. Acc. & $+0.000$ & $[+0.000, +0.000]$ & $1.000$ \\
     &  & ROC AUC & $+0.002$ & $[-0.006, +0.010]$ & $0.621$ \\
     &  & PR AUC & $+0.001$ & $[-0.006, +0.009]$ & $0.664$ \\
    \cmidrule(lr){2-6}
     & \multirow{3}{*}{TFM} & Bal. Acc. & $+0.067$ & $[+0.059, +0.074]$ & $<0.001$\textsuperscript{***} \\
     &  & ROC AUC & $-0.141$ & $[-0.166, -0.116]$ & $<0.001$\textsuperscript{***} \\
     &  & PR AUC & $-0.078$ & $[-0.101, -0.055]$ & $0.002$\textsuperscript{**} \\
    \cmidrule(lr){2-6}
     & \multirow{3}{*}{REVE-Base} & Bal. Acc. & $+0.189$ & $[+0.125, +0.253]$ & $0.002$\textsuperscript{**} \\
     &  & ROC AUC & $+0.011$ & $[+0.005, +0.017]$ & $0.014$\textsuperscript{*} \\
     &  & PR AUC & $-0.007$ & $[-0.044, +0.029]$ & $0.642$ \\
     & \multirow{3}{*}{REVE-Large} & Bal. Acc. & $+0.187$ & $[+0.140, +0.233]$ & $<0.001$\textsuperscript{***} \\
     &  & ROC AUC & $-0.001$ & $[-0.011, +0.009]$ & $0.801$ \\
     &  & PR AUC & $-0.023$ & $[-0.110, +0.064]$ & $0.537$ \\
    \cmidrule(lr){1-6}
    \multirow{12}{*}{\vspace{-15pt}Tent} & \multirow{3}{*}{CBraMod} & Bal. Acc. & $+0.000$ & $[-0.000, +0.001]$ & $0.408$ \\
     &  & ROC AUC & $-0.257$ & $[-0.269, -0.246]$ & $<0.001$\textsuperscript{***} \\
     &  & PR AUC & $-0.070$ & $[-0.105, -0.034]$ & $0.009$\textsuperscript{**} \\
    \cmidrule(lr){2-6}
     & \multirow{3}{*}{TFM} & Bal. Acc. & $-0.004$ & $[-0.009, +0.001]$ & $0.112$ \\
     &  & ROC AUC & $-0.001$ & $[-0.003, +0.000]$ & $0.142$ \\
     &  & PR AUC & $+0.020$ & $[+0.008, +0.033]$ & $0.017$\textsuperscript{*} \\
    \cmidrule(lr){2-6}
     & \multirow{3}{*}{REVE-Base} & Bal. Acc. & $-0.013$ & $[-0.034, +0.008]$ & $0.192$ \\
     &  & ROC AUC & $-0.002$ & $[-0.005, +0.001]$ & $0.127$ \\
     &  & PR AUC & $+0.009$ & $[+0.001, +0.018]$ & $0.048$\textsuperscript{*} \\
     & \multirow{3}{*}{REVE-Large} & Bal. Acc. & $-0.063$ & $[-0.094, -0.031]$ & $0.009$\textsuperscript{**} \\
     &  & ROC AUC & $-0.034$ & $[-0.041, -0.027]$ & $<0.001$\textsuperscript{***} \\
     &  & PR AUC & $-0.005$ & $[-0.023, +0.013]$ & $0.502$ \\
    \bottomrule
  \end{tabular}
  \end{adjustbox}
\end{table}

\begin{table}[H]
  \centering
  \scriptsize
  \setlength{\tabcolsep}{3pt}
  \renewcommand{\arraystretch}{1.05}
  \caption{Sleep-EDF paired seed-level $t$-tests comparing each TTA method against No-TTA, with Benjamini--Hochberg FDR correction. Asterisks denote FDR-adjusted significance levels, with \textsuperscript{*}$q<0.05$, \textsuperscript{**}$q<0.01$, and \textsuperscript{***}$q<0.001$. Unmarked comparisons are not significant.}
  \label{tab_sleep_edf_78_significance}
  \begin{adjustbox}{max width=\textwidth}
  \begin{tabular}{p{1.35cm}p{1.55cm}p{1.75cm}ccc}
    \toprule
    \textbf{\shortstack{TTA\\Method}} & \textbf{\shortstack{Foundation\\Model}} & \textbf{Metric} & \textbf{Mean $\Delta$} & \textbf{$95\%$ CI} & \textbf{$q$} \\
    \midrule
    \multirow{12}{*}{\vspace{-15pt}SHOT} & \multirow{3}{*}{CBraMod} & Bal. Acc. & $-0.315$ & $[-0.332, -0.298]$ & $<0.001$\textsuperscript{***} \\
     &  & Cohen's $\kappa$ & $-0.554$ & $[-0.558, -0.551]$ & $<0.001$\textsuperscript{***} \\
     &  & Weighted F1 & $-0.627$ & $[-0.676, -0.578]$ & $<0.001$\textsuperscript{***} \\
    \cmidrule(lr){2-6}
     & \multirow{3}{*}{TFM} & Bal. Acc. & $+0.004$ & $[+0.001, +0.008]$ & $0.026$\textsuperscript{*} \\
     &  & Cohen's $\kappa$ & $-0.006$ & $[-0.010, -0.001]$ & $0.026$\textsuperscript{*} \\
     &  & Weighted F1 & $-0.002$ & $[-0.004, -0.000]$ & $0.032$\textsuperscript{*} \\
    \cmidrule(lr){2-6}
     & \multirow{3}{*}{REVE-Base} & Bal. Acc. & $-0.315$ & $[-0.330, -0.301]$ & $<0.001$\textsuperscript{***} \\
     &  & Cohen's $\kappa$ & $-0.511$ & $[-0.541, -0.480]$ & $<0.001$\textsuperscript{***} \\
     &  & Weighted F1 & $-0.461$ & $[-0.491, -0.432]$ & $<0.001$\textsuperscript{***} \\
     & \multirow{3}{*}{REVE-Large} & Bal. Acc. & $-0.418$ & $[-0.435, -0.401]$ & $<0.001$\textsuperscript{***} \\
     &  & Cohen's $\kappa$ & $-0.640$ & $[-0.668, -0.613]$ & $<0.001$\textsuperscript{***} \\
     &  & Weighted F1 & $-0.664$ & $[-0.725, -0.603]$ & $<0.001$\textsuperscript{***} \\
    \cmidrule(lr){1-6}
    \multirow{12}{*}{\vspace{-15pt}T3A} & \multirow{3}{*}{CBraMod} & Bal. Acc. & $+0.022$ & $[+0.009, +0.034]$ & $0.012$\textsuperscript{*} \\
     &  & Cohen's $\kappa$ & $-0.154$ & $[-0.171, -0.137]$ & $<0.001$\textsuperscript{***} \\
     &  & Weighted F1 & $-0.093$ & $[-0.109, -0.076]$ & $<0.001$\textsuperscript{***} \\
    \cmidrule(lr){2-6}
     & \multirow{3}{*}{TFM} & Bal. Acc. & $-0.014$ & $[-0.023, -0.004]$ & $0.024$\textsuperscript{*} \\
     &  & Cohen's $\kappa$ & $-0.130$ & $[-0.157, -0.103]$ & $<0.001$\textsuperscript{***} \\
     &  & Weighted F1 & $-0.076$ & $[-0.098, -0.054]$ & $0.002$\textsuperscript{**} \\
    \cmidrule(lr){2-6}
     & \multirow{3}{*}{REVE-Base} & Bal. Acc. & $-0.042$ & $[-0.055, -0.029]$ & $0.002$\textsuperscript{**} \\
     &  & Cohen's $\kappa$ & $-0.229$ & $[-0.248, -0.210]$ & $<0.001$\textsuperscript{***} \\
     &  & Weighted F1 & $-0.160$ & $[-0.180, -0.141]$ & $<0.001$\textsuperscript{***} \\
     & \multirow{3}{*}{REVE-Large} & Bal. Acc. & $-0.026$ & $[-0.034, -0.018]$ & $0.002$\textsuperscript{**} \\
     &  & Cohen's $\kappa$ & $-0.215$ & $[-0.235, -0.194]$ & $<0.001$\textsuperscript{***} \\
     &  & Weighted F1 & $-0.144$ & $[-0.163, -0.124]$ & $<0.001$\textsuperscript{***} \\
    \cmidrule(lr){1-6}
    \multirow{12}{*}{\vspace{-15pt}Tent} & \multirow{3}{*}{CBraMod} & Bal. Acc. & $-0.312$ & $[-0.330, -0.294]$ & $<0.001$\textsuperscript{***} \\
     &  & Cohen's $\kappa$ & $-0.552$ & $[-0.556, -0.548]$ & $<0.001$\textsuperscript{***} \\
     &  & Weighted F1 & $-0.508$ & $[-0.532, -0.483]$ & $<0.001$\textsuperscript{***} \\
    \cmidrule(lr){2-6}
     & \multirow{3}{*}{TFM} & Bal. Acc. & $-0.076$ & $[-0.111, -0.042]$ & $0.006$\textsuperscript{**} \\
     &  & Cohen's $\kappa$ & $-0.086$ & $[-0.153, -0.020]$ & $0.030$\textsuperscript{*} \\
     &  & Weighted F1 & $-0.090$ & $[-0.143, -0.037]$ & $0.014$\textsuperscript{*} \\
    \cmidrule(lr){2-6}
     & \multirow{3}{*}{REVE-Base} & Bal. Acc. & $-0.183$ & $[-0.226, -0.140]$ & $<0.001$\textsuperscript{***} \\
     &  & Cohen's $\kappa$ & $-0.176$ & $[-0.278, -0.074]$ & $0.013$\textsuperscript{*} \\
     &  & Weighted F1 & $-0.160$ & $[-0.263, -0.057]$ & $0.018$\textsuperscript{*} \\
     & \multirow{3}{*}{REVE-Large} & Bal. Acc. & $-0.155$ & $[-0.162, -0.147]$ & $<0.001$\textsuperscript{***} \\
     &  & Cohen's $\kappa$ & $-0.121$ & $[-0.154, -0.088]$ & $0.001$\textsuperscript{**} \\
     &  & Weighted F1 & $-0.102$ & $[-0.121, -0.084]$ & $<0.001$\textsuperscript{***} \\
    \bottomrule
  \end{tabular}
  \end{adjustbox}
\end{table}

\begin{table}[H]
  \centering
  \scriptsize
  \setlength{\tabcolsep}{3pt}
  \renewcommand{\arraystretch}{1.05}
  \caption{EAR-EEG paired seed-level $t$-tests comparing each TTA method against No-TTA, with Benjamini--Hochberg FDR correction. Asterisks denote FDR-adjusted significance levels, with \textsuperscript{*}$q<0.05$, \textsuperscript{**}$q<0.01$, and \textsuperscript{***}$q<0.001$. Unmarked comparisons are not significant.}
  \label{tab_eareeg_significance}
  \begin{adjustbox}{max width=\textwidth}
  \begin{tabular}{p{1.35cm}p{1.55cm}p{1.75cm}ccc}
    \toprule
    \textbf{\shortstack{TTA\\Method}} & \textbf{\shortstack{Foundation\\Model}} & \textbf{Metric} & \textbf{Mean $\Delta$} & \textbf{$95\%$ CI} & \textbf{$q$} \\
    \midrule
    \multirow{12}{*}{\vspace{-15pt}SHOT} & \multirow{3}{*}{CBraMod} & Bal. Acc. & $-0.065$ & $[-0.089, -0.042]$ & $0.003$\textsuperscript{**} \\
     &  & Cohen's $\kappa$ & $-0.087$ & $[-0.119, -0.054]$ & $0.003$\textsuperscript{**} \\
     &  & Weighted F1 & $-0.224$ & $[-0.238, -0.210]$ & $<0.001$\textsuperscript{***} \\
    \cmidrule(lr){2-6}
     & \multirow{3}{*}{TFM} & Bal. Acc. & $+0.000$ & $[-0.000, +0.000]$ & $0.777$ \\
     &  & Cohen's $\kappa$ & $-0.000$ & $[-0.001, +0.000]$ & $0.514$ \\
     &  & Weighted F1 & $-0.000$ & $[-0.001, +0.000]$ & $0.204$ \\
    \cmidrule(lr){2-6}
     & \multirow{3}{*}{REVE-Base} & Bal. Acc. & $-0.018$ & $[-0.031, -0.006]$ & $0.021$\textsuperscript{*} \\
     &  & Cohen's $\kappa$ & $-0.085$ & $[-0.099, -0.071]$ & $<0.001$\textsuperscript{***} \\
     &  & Weighted F1 & $-0.119$ & $[-0.133, -0.105]$ & $<0.001$\textsuperscript{***} \\
     & \multirow{3}{*}{REVE-Large} & Bal. Acc. & $-0.068$ & $[-0.118, -0.018]$ & $0.026$\textsuperscript{*} \\
     &  & Cohen's $\kappa$ & $-0.156$ & $[-0.221, -0.090]$ & $0.005$\textsuperscript{**} \\
     &  & Weighted F1 & $-0.150$ & $[-0.207, -0.094]$ & $0.003$\textsuperscript{**} \\
    \cmidrule(lr){1-6}
    \multirow{12}{*}{\vspace{-15pt}T3A} & \multirow{3}{*}{CBraMod} & Bal. Acc. & $+0.048$ & $[+0.033, +0.063]$ & $0.002$\textsuperscript{**} \\
     &  & Cohen's $\kappa$ & $+0.064$ & $[+0.043, +0.085]$ & $0.002$\textsuperscript{**} \\
     &  & Weighted F1 & $+0.018$ & $[+0.006, +0.029]$ & $0.018$\textsuperscript{*} \\
    \cmidrule(lr){2-6}
     & \multirow{3}{*}{TFM} & Bal. Acc. & $-0.005$ & $[-0.014, +0.005]$ & $0.279$ \\
     &  & Cohen's $\kappa$ & $-0.042$ & $[-0.050, -0.034]$ & $<0.001$\textsuperscript{***} \\
     &  & Weighted F1 & $-0.009$ & $[-0.016, -0.002]$ & $0.031$\textsuperscript{*} \\
    \cmidrule(lr){2-6}
     & \multirow{3}{*}{REVE-Base} & Bal. Acc. & $+0.037$ & $[+0.028, +0.045]$ & $<0.001$\textsuperscript{***} \\
     &  & Cohen's $\kappa$ & $+0.001$ & $[-0.018, +0.020]$ & $0.903$ \\
     &  & Weighted F1 & $+0.001$ & $[-0.022, +0.023]$ & $0.931$ \\
     & \multirow{3}{*}{REVE-Large} & Bal. Acc. & $+0.022$ & $[+0.014, +0.031]$ & $0.004$\textsuperscript{**} \\
     &  & Cohen's $\kappa$ & $-0.010$ & $[-0.028, +0.009]$ & $0.243$ \\
     &  & Weighted F1 & $-0.009$ & $[-0.023, +0.004]$ & $0.150$ \\
    \cmidrule(lr){1-6}
    \multirow{12}{*}{\vspace{-15pt}Tent} & \multirow{3}{*}{CBraMod} & Bal. Acc. & $-0.064$ & $[-0.085, -0.043]$ & $0.002$\textsuperscript{**} \\
     &  & Cohen's $\kappa$ & $-0.084$ & $[-0.116, -0.052]$ & $0.003$\textsuperscript{**} \\
     &  & Weighted F1 & $-0.189$ & $[-0.226, -0.153]$ & $<0.001$\textsuperscript{***} \\
    \cmidrule(lr){2-6}
     & \multirow{3}{*}{TFM} & Bal. Acc. & $-0.001$ & $[-0.001, +0.000]$ & $0.197$ \\
     &  & Cohen's $\kappa$ & $+0.000$ & $[-0.000, +0.001]$ & $0.267$ \\
     &  & Weighted F1 & $-0.000$ & $[-0.001, +0.000]$ & $0.072$ \\
    \cmidrule(lr){2-6}
     & \multirow{3}{*}{REVE-Base} & Bal. Acc. & $-0.032$ & $[-0.042, -0.022]$ & $0.002$\textsuperscript{**} \\
     &  & Cohen's $\kappa$ & $-0.047$ & $[-0.062, -0.031]$ & $0.002$\textsuperscript{**} \\
     &  & Weighted F1 & $-0.046$ & $[-0.059, -0.033]$ & $0.002$\textsuperscript{**} \\
     & \multirow{3}{*}{REVE-Large} & Bal. Acc. & $-0.018$ & $[-0.024, -0.012]$ & $0.003$\textsuperscript{**} \\
     &  & Cohen's $\kappa$ & $-0.025$ & $[-0.033, -0.016]$ & $0.003$\textsuperscript{**} \\
     &  & Weighted F1 & $-0.019$ & $[-0.028, -0.010]$ & $0.007$\textsuperscript{**} \\
    \bottomrule
  \end{tabular}
  \end{adjustbox}
\end{table}

\subsection{Test-Time Computational Cost}
\label{sec:appendix_compute_cost}

We report measured adaptation cost on the full evaluation grid. Runtime is the
wall-clock time spent in the adaptation step before final evaluation. Peak
memory is the maximum CUDA allocation observed during adaptation. No-TTA has no
adaptation step and is not included in the table.
Each entry aggregates 75 runs across five datasets (TUEV, TUAB, CHB-MIT,
Sleep-EDF, EAR-EEG), five seeds (3, 5, 6, 5635, 10996), and three adaptation
batch sizes (64, 128, 256). Table~\ref{tab:compute_cost_summary} reports adaptation latency and memory
requirements across all datasets, seeds, and adaptation batch sizes.

\begin{table}[H]
  \centering
  \scriptsize
  \setlength{\tabcolsep}{3pt}
  \renewcommand{\arraystretch}{1.05}
  \caption{Measured adaptation cost on NVIDIA H200 GPUs, summarized by TTA method and foundation model. Runtime is reported per EEG window and peak memory is the maximum CUDA allocation observed during adaptation.}
  \label{tab:compute_cost_summary}
  \begin{adjustbox}{max width=\textwidth}
  \begin{tabular}{p{1.35cm}p{1.8cm}p{1.65cm}p{1.45cm}p{1.50cm}p{1.35cm}}
    \toprule
    \textbf{\shortstack{TTA\\Method}} & \textbf{\shortstack{Foundation\\Model}} & \textbf{\shortstack{Median\\$\mu\mathrm{s/window}$}} & \textbf{\shortstack{Max\\$\mu\mathrm{s/window}$}} & \textbf{\shortstack{Median\\peak GB}} & \textbf{\shortstack{Max\\peak GB}} \\
    \midrule
    \multirow{4}{*}{\vspace{-10pt}T3A} & CBraMod & 0.16 & 1.89 & 0.08 & 0.08 \\
    \cmidrule(lr){2-6}
     & TFM & 0.28 & 3.41 & 0.07 & 0.07 \\
    \cmidrule(lr){2-6}
     & REVE-Base & 0.25 & 2.97 & 0.32 & 0.32 \\
     & REVE-Large & 0.25 & 2.74 & 1.52 & 1.52 \\
    \cmidrule(lr){1-6}
    \multirow{4}{*}{\vspace{-10pt}Tent} & CBraMod & 0.26 & 2.68 & 0.07 & 0.07 \\
    \cmidrule(lr){2-6}
     & TFM & 0.26 & 2.70 & 0.04 & 0.04 \\
    \cmidrule(lr){2-6}
     & REVE-Base & 0.28 & 3.52 & 0.55 & 0.55 \\
     & REVE-Large & 0.39 & 5.36 & 2.94 & 2.94 \\
    \cmidrule(lr){1-6}
    \multirow{4}{*}{\vspace{-10pt}SHOT} & CBraMod & 190.82 & 2496.89 & 0.25 & 0.54 \\
    \cmidrule(lr){2-6}
     & TFM & 285.94 & 3349.99 & 1.61 & 6.22 \\
    \cmidrule(lr){2-6}
     & REVE-Base & 768.53 & 3403.66 & 0.96 & 2.11 \\
     & REVE-Large & 3183.54 & 4305.95 & 3.86 & 6.55 \\
    \bottomrule
  \end{tabular}
  \end{adjustbox}
\end{table}

\subsection{Source-Target Structure Visualization}
\label{sec:appendix_visualization}

We use t-SNE to qualitatively inspect source-target structure under the
distribution shifts evaluated in the benchmark. We use source for the training
split and target for the held-out test split. For each dataset, the first column
shows raw EEG and the remaining columns show frozen foundation-model embeddings
before the downstream classifier. The top row colors points by task class, and
the bottom row colors the same points by source or target split. Visible
source-target structure suggests that the representation is not fully
domain-invariant in this diagnostic view. These plots motivate evaluating
test-time adaptation, while quantitative conclusions come from the benchmark
results and significance analyses.
For EAR-EEG, this diagnostic is especially relevant because it is an unseen
ear-EEG modality, while the evaluated foundation models were pretrained
primarily on scalp EEG.

\begin{figure}[H]
  \centering
  \includegraphics[width=\linewidth]{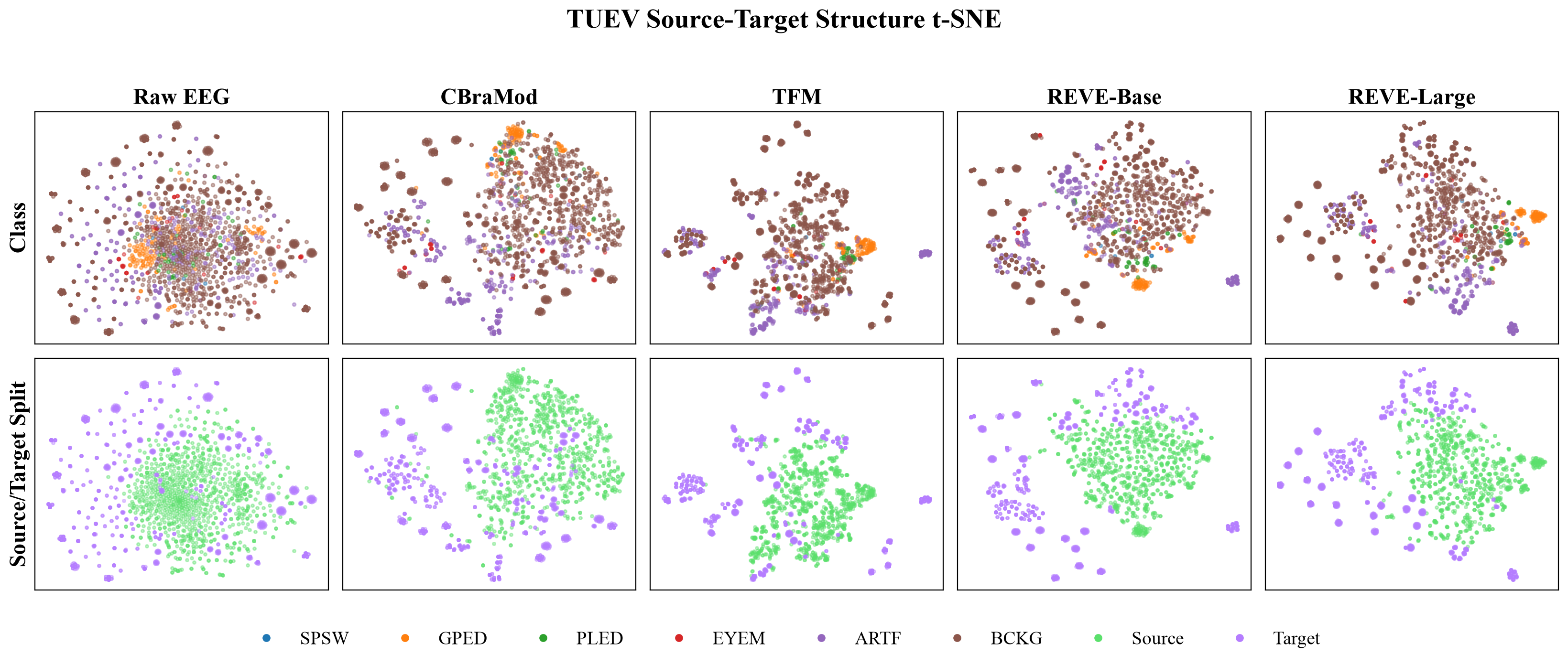}
  \caption{TUEV source-target t-SNE visualization for raw EEG and frozen
  foundation-model embeddings.}
  \label{fig:embedding_shift_tuev}
\end{figure}

\begin{figure}[H]
  \centering
  \includegraphics[width=\linewidth]{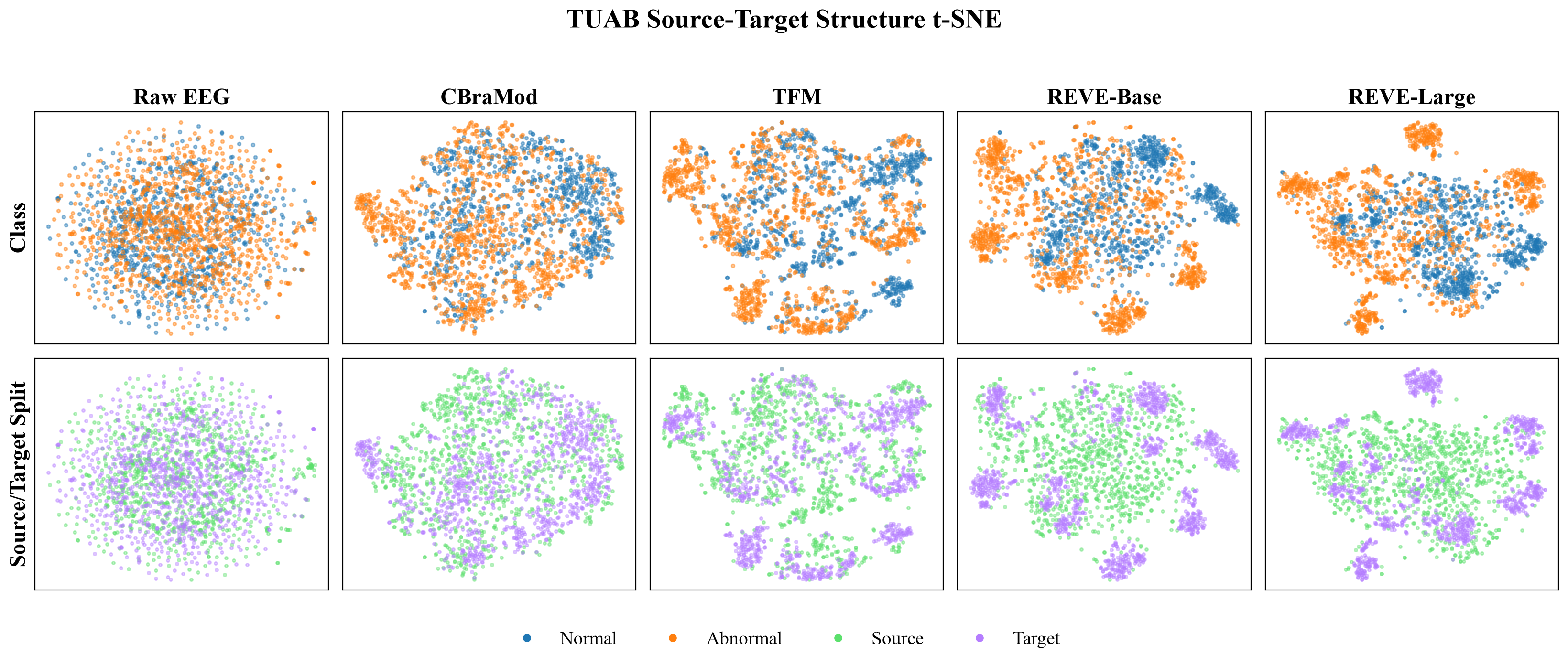}
  \caption{TUAB source-target t-SNE visualization for raw EEG and frozen
  foundation-model embeddings.}
  \label{fig:embedding_shift_tuab}
\end{figure}

\begin{figure}[H]
  \centering
  \includegraphics[width=\linewidth]{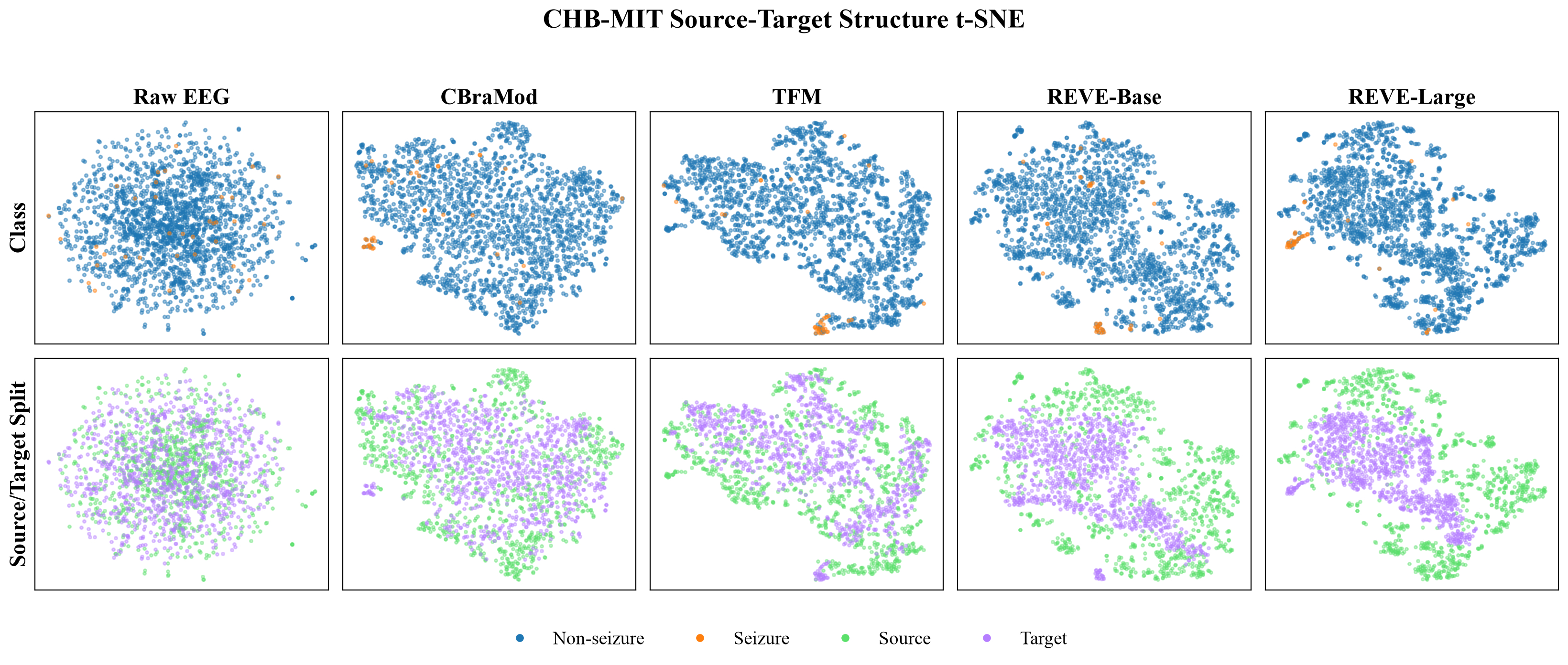}
  \caption{CHB-MIT source-target t-SNE visualization for raw EEG and frozen
  foundation-model embeddings.}
  \label{fig:embedding_shift_chbmit}
\end{figure}

\begin{figure}[H]
  \centering
  \includegraphics[width=\linewidth]{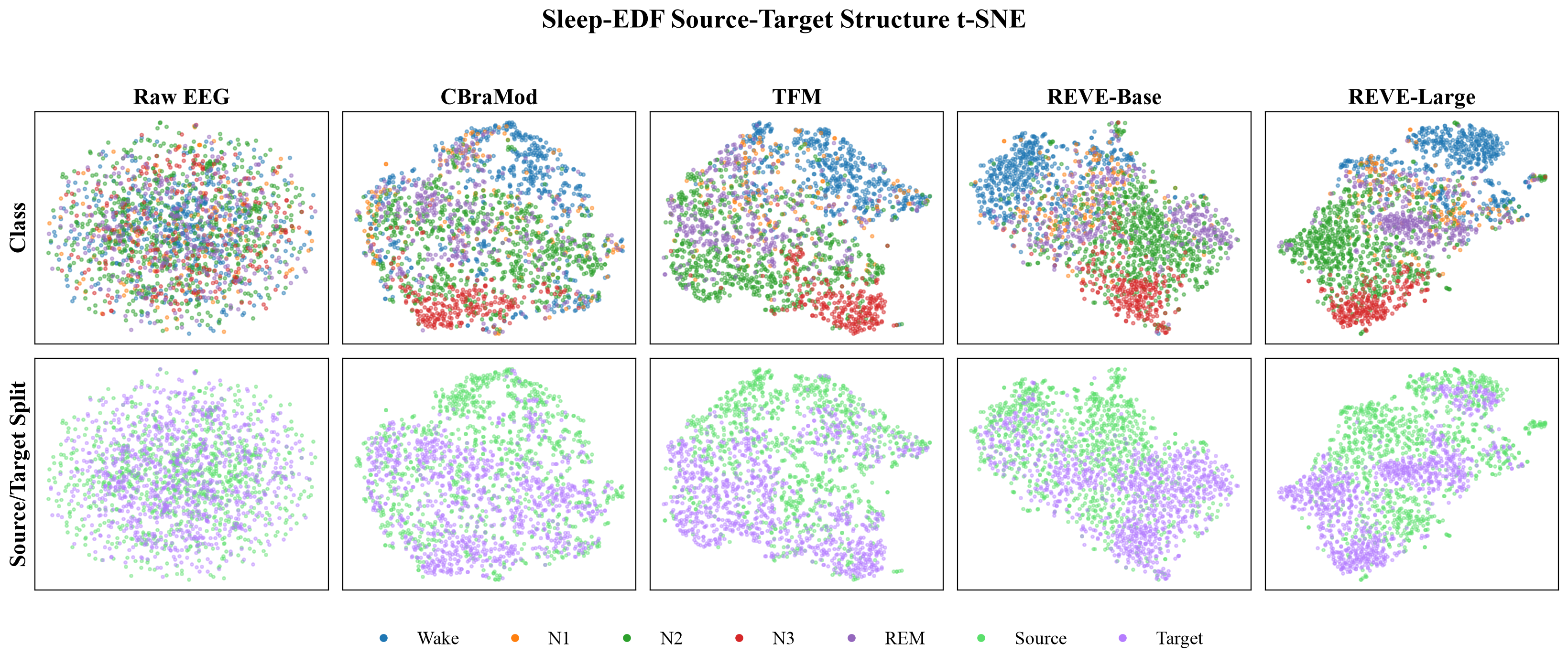}
  \caption{Sleep-EDF source-target t-SNE visualization for raw EEG and frozen
  foundation-model embeddings.}
  \label{fig:embedding_shift_sleep_edf_78}
\end{figure}

\begin{figure}[H]
  \centering
  \includegraphics[width=\linewidth]{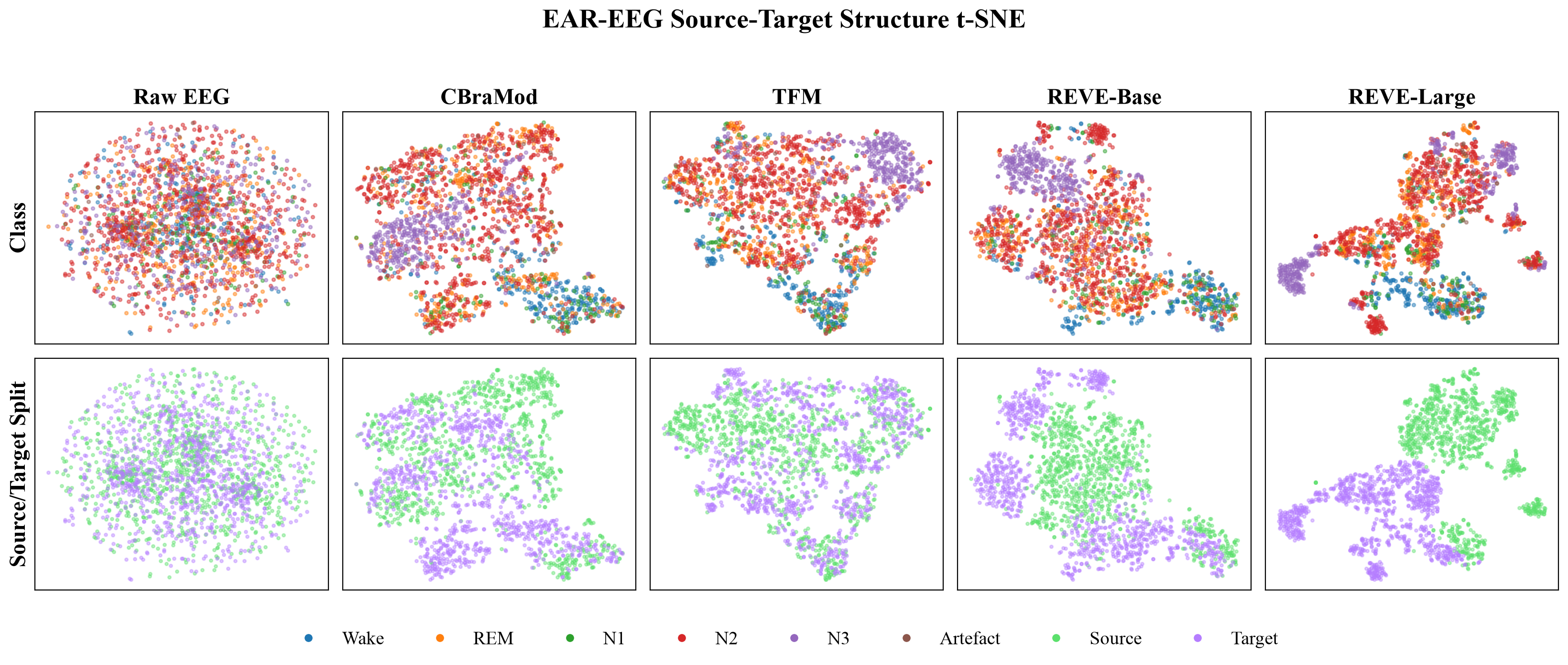}
  \caption{EAR-EEG source-target t-SNE visualization for raw EEG and frozen
  foundation-model embeddings.}
  \label{fig:embedding_shift_eareeg}
\end{figure}

\subsection{Absolute Performance by Dataset}
Values are mean $\pm$ standard deviation aggregated across adaptation batch sizes ($64$, $128$, $256$) and reported study seeds.

\begin{table}[H]
\centering
\caption{CHB-MIT and TUAB}
\fontsize{8}{5}\selectfont
\setlength{\tabcolsep}{3pt}
\renewcommand{\arraystretch}{1.05}
\begin{tabular}{p{1.75cm}p{1.4cm}p{2.2cm}p{2.2cm}p{2.2cm}p{2.2cm}}
\toprule
\multirow{2}{*}{\textbf{\vspace{-3pt}Dataset}} & \multirow{2}{*}{\textbf{\shortstack{\vspace{-3pt}Method}}} & \multirow{2}{*}{\textbf{\shortstack{Foundation\\Model}}} & \multicolumn{3}{c}{\textbf{Performance Metrics}} \\
\cmidrule(lr){4-6}
& & & \textbf{Bal. Acc.} & \textbf{PR AUC} & \textbf{ROC AUC} \\
\midrule
\multirow[c]{17}{*}{\vspace{-75pt}CHB-MIT}
& \multirow{4}{*}{\vspace{-15pt}No-TTA} & CBraMod & $0.500 \pm 0.000$ & $0.093 \pm 0.015$ & $0.750 \pm 0.020$ \\
\cmidrule(lr){3-6}
&  & TFM & $0.534 \pm 0.006$ & $0.331 \pm 0.024$ & $0.864 \pm 0.007$ \\
\cmidrule(lr){3-6}
&  & REVE-Base & $0.552 \pm 0.039$ & $0.318 \pm 0.019$ & $0.843 \pm 0.010$ \\
&  & REVE-Large & $0.608 \pm 0.042$ & $0.404 \pm 0.020$ & $0.890 \pm 0.015$ \\
\cmidrule(lr){2-6}
& \multirow{4}{*}{\vspace{-15pt}SHOT} & CBraMod & $0.514 \pm 0.030$ & $0.024 \pm 0.013$ & $0.504 \pm 0.025$ \\
\cmidrule(lr){3-6}
&  & TFM & $0.541 \pm 0.012$ & $0.304 \pm 0.038$ & $0.861 \pm 0.011$ \\
\cmidrule(lr){3-6}
&  & REVE-Base & $0.556 \pm 0.070$ & $0.031 \pm 0.010$ & $0.518 \pm 0.131$ \\
&  & REVE-Large & $0.590 \pm 0.074$ & $0.072 \pm 0.091$ & $0.680 \pm 0.126$ \\
\cmidrule(lr){2-6}
& \multirow{4}{*}{\vspace{-15pt}Tent} & CBraMod & $0.500 \pm 0.001$ & $0.023 \pm 0.018$ & $0.493 \pm 0.057$ \\
\cmidrule(lr){3-6}
&  & TFM & $0.530 \pm 0.003$ & $0.351 \pm 0.022$ & $0.863 \pm 0.006$ \\
\cmidrule(lr){3-6}
&  & REVE-Base & $0.539 \pm 0.025$ & $0.327 \pm 0.016$ & $0.840 \pm 0.010$ \\
&  & REVE-Large & $0.545 \pm 0.026$ & $0.398 \pm 0.028$ & $0.856 \pm 0.015$ \\
\cmidrule(lr){2-6}
& \multirow{4}{*}{\vspace{-15pt}T3A} & CBraMod & $0.500 \pm 0.000$ & $0.094 \pm 0.019$ & $0.752 \pm 0.024$ \\
\cmidrule(lr){3-6}
&  & TFM & $0.601 \pm 0.011$ & $0.253 \pm 0.011$ & $0.723 \pm 0.020$ \\
\cmidrule(lr){3-6}
&  & REVE-Base & $0.741 \pm 0.019$ & $0.310 \pm 0.027$ & $0.853 \pm 0.012$ \\
&  & REVE-Large & $0.795 \pm 0.010$ & $0.380 \pm 0.054$ & $0.889 \pm 0.010$ \\
\midrule
\multirow[c]{17}{*}{\vspace{-75pt}TUAB}
& \multirow{4}{*}{\vspace{-15pt}No-TTA} & CBraMod & $0.749 \pm 0.004$ & $0.823 \pm 0.002$ & $0.822 \pm 0.001$ \\
\cmidrule(lr){3-6}
&  & TFM & $0.761 \pm 0.007$ & $0.830 \pm 0.003$ & $0.844 \pm 0.004$ \\
\cmidrule(lr){3-6}
&  & REVE-Base & $0.801 \pm 0.005$ & $0.889 \pm 0.004$ & $0.879 \pm 0.004$ \\
&  & REVE-Large & $0.810 \pm 0.005$ & $0.899 \pm 0.003$ & $0.889 \pm 0.004$ \\
\cmidrule(lr){2-6}
& \multirow{4}{*}{\vspace{-15pt}SHOT} & CBraMod & $0.502 \pm 0.004$ & $0.486 \pm 0.032$ & $0.519 \pm 0.049$ \\
\cmidrule(lr){3-6}
&  & TFM & $0.732 \pm 0.014$ & $0.818 \pm 0.007$ & $0.832 \pm 0.009$ \\
\cmidrule(lr){3-6}
&  & REVE-Base & $0.654 \pm 0.058$ & $0.791 \pm 0.065$ & $0.790 \pm 0.059$ \\
&  & REVE-Large & $0.700 \pm 0.057$ & $0.839 \pm 0.033$ & $0.833 \pm 0.032$ \\
\cmidrule(lr){2-6}
& \multirow{4}{*}{\vspace{-15pt}Tent} & CBraMod & $0.501 \pm 0.003$ & $0.499 \pm 0.028$ & $0.551 \pm 0.041$ \\
\cmidrule(lr){3-6}
&  & TFM & $0.565 \pm 0.035$ & $0.656 \pm 0.041$ & $0.661 \pm 0.032$ \\
\cmidrule(lr){3-6}
&  & REVE-Base & $0.578 \pm 0.044$ & $0.645 \pm 0.074$ & $0.656 \pm 0.059$ \\
&  & REVE-Large & $0.644 \pm 0.082$ & $0.751 \pm 0.088$ & $0.755 \pm 0.076$ \\
\cmidrule(lr){2-6}
& \multirow{4}{*}{\vspace{-15pt}T3A} & CBraMod & $0.724 \pm 0.006$ & $0.794 \pm 0.006$ & $0.801 \pm 0.004$ \\
\cmidrule(lr){3-6}
&  & TFM & $0.734 \pm 0.003$ & $0.788 \pm 0.004$ & $0.813 \pm 0.004$ \\
\cmidrule(lr){3-6}
&  & REVE-Base & $0.787 \pm 0.005$ & $0.873 \pm 0.002$ & $0.867 \pm 0.002$ \\
&  & REVE-Large & $0.807 \pm 0.005$ & $0.892 \pm 0.004$ & $0.885 \pm 0.007$ \\
\bottomrule
\end{tabular}
\end{table}

\vspace{0.5em}

\begin{table}[H]
\centering
\caption{EarEEG, SleepEDF-78, and TUEV}
\fontsize{8}{5}\selectfont
\setlength{\tabcolsep}{3pt}
\renewcommand{\arraystretch}{1.05}
\begin{tabular}{p{1.75cm}p{1.4cm}p{2.2cm}p{2.2cm}p{2.2cm}p{2.2cm}}
\toprule
\multirow{2}{*}{\textbf{\vspace{-3pt}Dataset}} & \multirow{2}{*}{\textbf{\shortstack{\vspace{-3pt}Method}}} & \multirow{2}{*}{\textbf{\shortstack{Foundation\\Model}}} & \multicolumn{3}{c}{\textbf{Performance Metrics}} \\
\cmidrule(lr){4-6}
& & & \textbf{Bal. Acc.} & \textbf{Cohen's $\kappa$} & \textbf{Weighted F1} \\
\midrule
\multirow[c]{17}{*}{\vspace{-75pt}EarEEG}
& \multirow{4}{*}{\vspace{-15pt}No-TTA} & CBraMod & $0.238 \pm 0.019$ & $0.093 \pm 0.026$ & $0.299 \pm 0.012$ \\
\cmidrule(lr){3-6}
&  & TFM & $0.371 \pm 0.007$ & $0.285 \pm 0.007$ & $0.426 \pm 0.005$ \\
\cmidrule(lr){3-6}
&  & REVE-Base & $0.360 \pm 0.008$ & $0.307 \pm 0.010$ & $0.468 \pm 0.008$ \\
&  & REVE-Large & $0.400 \pm 0.018$ & $0.373 \pm 0.021$ & $0.510 \pm 0.013$ \\
\cmidrule(lr){2-6}
& \multirow{4}{*}{\vspace{-15pt}SHOT} & CBraMod & $0.173 \pm 0.012$ & $0.007 \pm 0.013$ & $0.075 \pm 0.015$ \\
\cmidrule(lr){3-6}
&  & TFM & $0.372 \pm 0.007$ & $0.285 \pm 0.007$ & $0.426 \pm 0.005$ \\
\cmidrule(lr){3-6}
&  & REVE-Base & $0.342 \pm 0.034$ & $0.222 \pm 0.065$ & $0.349 \pm 0.076$ \\
&  & REVE-Large & $0.332 \pm 0.064$ & $0.218 \pm 0.091$ & $0.360 \pm 0.091$ \\
\cmidrule(lr){2-6}
& \multirow{4}{*}{\vspace{-15pt}Tent} & CBraMod & $0.174 \pm 0.018$ & $0.009 \pm 0.021$ & $0.109 \pm 0.057$ \\
\cmidrule(lr){3-6}
&  & TFM & $0.371 \pm 0.007$ & $0.286 \pm 0.007$ & $0.425 \pm 0.005$ \\
\cmidrule(lr){3-6}
&  & REVE-Base & $0.328 \pm 0.022$ & $0.260 \pm 0.035$ & $0.422 \pm 0.029$ \\
&  & REVE-Large & $0.382 \pm 0.023$ & $0.348 \pm 0.030$ & $0.491 \pm 0.022$ \\
\cmidrule(lr){2-6}
& \multirow{4}{*}{\vspace{-15pt}T3A} & CBraMod & $0.286 \pm 0.012$ & $0.157 \pm 0.016$ & $0.316 \pm 0.006$ \\
\cmidrule(lr){3-6}
&  & TFM & $0.367 \pm 0.009$ & $0.243 \pm 0.007$ & $0.417 \pm 0.003$ \\
\cmidrule(lr){3-6}
&  & REVE-Base & $0.397 \pm 0.014$ & $0.308 \pm 0.020$ & $0.469 \pm 0.017$ \\
&  & REVE-Large & $0.422 \pm 0.014$ & $0.364 \pm 0.020$ & $0.501 \pm 0.013$ \\
\midrule
\multirow[c]{17}{*}{\vspace{-75pt}SleepEDF-78}
& \multirow{4}{*}{\vspace{-15pt}No-TTA} & CBraMod & $0.514 \pm 0.013$ & $0.554 \pm 0.002$ & $0.661 \pm 0.004$ \\
\cmidrule(lr){3-6}
&  & TFM & $0.564 \pm 0.003$ & $0.577 \pm 0.004$ & $0.686 \pm 0.003$ \\
\cmidrule(lr){3-6}
&  & REVE-Base & $0.622 \pm 0.006$ & $0.637 \pm 0.006$ & $0.735 \pm 0.003$ \\
&  & REVE-Large & $0.651 \pm 0.003$ & $0.678 \pm 0.004$ & $0.766 \pm 0.003$ \\
\cmidrule(lr){2-6}
& \multirow{4}{*}{\vspace{-15pt}SHOT} & CBraMod & $0.199 \pm 0.002$ & $-0.000 \pm 0.001$ & $0.034 \pm 0.043$ \\
\cmidrule(lr){3-6}
&  & TFM & $0.568 \pm 0.004$ & $0.572 \pm 0.005$ & $0.684 \pm 0.003$ \\
\cmidrule(lr){3-6}
&  & REVE-Base & $0.307 \pm 0.061$ & $0.126 \pm 0.075$ & $0.273 \pm 0.101$ \\
&  & REVE-Large & $0.233 \pm 0.021$ & $0.038 \pm 0.026$ & $0.102 \pm 0.059$ \\
\cmidrule(lr){2-6}
& \multirow{4}{*}{\vspace{-15pt}Tent} & CBraMod & $0.202 \pm 0.002$ & $0.002 \pm 0.002$ & $0.153 \pm 0.039$ \\
\cmidrule(lr){3-6}
&  & TFM & $0.488 \pm 0.058$ & $0.491 \pm 0.109$ & $0.596 \pm 0.087$ \\
\cmidrule(lr){3-6}
&  & REVE-Base & $0.439 \pm 0.091$ & $0.461 \pm 0.175$ & $0.575 \pm 0.173$ \\
&  & REVE-Large & $0.496 \pm 0.052$ & $0.557 \pm 0.082$ & $0.664 \pm 0.055$ \\
\cmidrule(lr){2-6}
& \multirow{4}{*}{\vspace{-15pt}T3A} & CBraMod & $0.535 \pm 0.008$ & $0.400 \pm 0.011$ & $0.568 \pm 0.010$ \\
\cmidrule(lr){3-6}
&  & TFM & $0.550 \pm 0.006$ & $0.447 \pm 0.018$ & $0.610 \pm 0.014$ \\
\cmidrule(lr){3-6}
&  & REVE-Base & $0.580 \pm 0.005$ & $0.408 \pm 0.012$ & $0.575 \pm 0.013$ \\
&  & REVE-Large & $0.625 \pm 0.006$ & $0.463 \pm 0.013$ & $0.622 \pm 0.013$ \\
\midrule
\multirow[c]{17}{*}{\vspace{-75pt}TUEV}
& \multirow{4}{*}{\vspace{-15pt}No-TTA} & CBraMod & $0.378 \pm 0.026$ & $0.464 \pm 0.056$ & $0.720 \pm 0.028$ \\
\cmidrule(lr){3-6}
&  & TFM & $0.369 \pm 0.011$ & $0.399 \pm 0.030$ & $0.684 \pm 0.016$ \\
\cmidrule(lr){3-6}
&  & REVE-Base & $0.454 \pm 0.016$ & $0.559 \pm 0.038$ & $0.771 \pm 0.018$ \\
&  & REVE-Large & $0.494 \pm 0.007$ & $0.585 \pm 0.018$ & $0.786 \pm 0.009$ \\
\cmidrule(lr){2-6}
& \multirow{4}{*}{\vspace{-15pt}SHOT} & CBraMod & $0.168 \pm 0.001$ & $0.003 \pm 0.002$ & $0.047 \pm 0.016$ \\
\cmidrule(lr){3-6}
&  & TFM & $0.394 \pm 0.017$ & $0.398 \pm 0.025$ & $0.674 \pm 0.019$ \\
\cmidrule(lr){3-6}
&  & REVE-Base & $0.359 \pm 0.099$ & $0.128 \pm 0.064$ & $0.197 \pm 0.086$ \\
&  & REVE-Large & $0.357 \pm 0.038$ & $0.112 \pm 0.031$ & $0.161 \pm 0.049$ \\
\cmidrule(lr){2-6}
& \multirow{4}{*}{\vspace{-15pt}Tent} & CBraMod & $0.166 \pm 0.003$ & $-0.001 \pm 0.002$ & $0.467 \pm 0.177$ \\
\cmidrule(lr){3-6}
&  & TFM & $0.268 \pm 0.039$ & $0.231 \pm 0.080$ & $0.616 \pm 0.029$ \\
\cmidrule(lr){3-6}
&  & REVE-Base & $0.204 \pm 0.025$ & $0.085 \pm 0.067$ & $0.572 \pm 0.029$ \\
&  & REVE-Large & $0.286 \pm 0.061$ & $0.258 \pm 0.121$ & $0.646 \pm 0.050$ \\
\cmidrule(lr){2-6}
& \multirow{4}{*}{\vspace{-15pt}T3A} & CBraMod & $0.408 \pm 0.032$ & $0.346 \pm 0.040$ & $0.624 \pm 0.037$ \\
\cmidrule(lr){3-6}
&  & TFM & $0.381 \pm 0.045$ & $0.219 \pm 0.026$ & $0.510 \pm 0.039$ \\
\cmidrule(lr){3-6}
&  & REVE-Base & $0.526 \pm 0.050$ & $0.422 \pm 0.106$ & $0.676 \pm 0.089$ \\
&  & REVE-Large & $0.588 \pm 0.048$ & $0.505 \pm 0.052$ & $0.742 \pm 0.031$ \\
\bottomrule
\end{tabular}
\end{table}

\end{document}